\documentclass[10pt,journal,compsoc]{IEEEtran}
\usepackage{amssymb}
\usepackage{booktabs}
\usepackage{multirow}
\usepackage[table]{xcolor}   
\usepackage{tabularx}
\usepackage{makecell}
\usepackage{threeparttable}
\usepackage[pagebackref,breaklinks,colorlinks]{hyperref}
\usepackage{algorithm}
\usepackage{algpseudocode}
\usepackage{etoolbox}

\makeatletter
\newcommand*{\algrule}[1][\algorithmicindent]{%
	\makebox[#1][l]{%
		\hspace*{.2em}
		\vrule height .75\baselineskip depth .25\baselineskip
	}
}

\newcount\ALG@printindent@tempcnta
\def\ALG@printindent{%
	\ifnum \theALG@nested>0
	\ifx\ALG@text\ALG@x@notext
	\else
	\unskip
	\ALG@printindent@tempcnta=1
	\loop
	\algrule[\csname ALG@ind@\the\ALG@printindent@tempcnta\endcsname]%
	\advance \ALG@printindent@tempcnta 1
	\ifnum \ALG@printindent@tempcnta<\numexpr\theALG@nested+1\relax
	\repeat
	\fi
	\fi
}
\patchcmd{\ALG@doentity}{\noindent\hskip\ALG@tlm}{\ALG@printindent}{}{\errmessage{failed to patch}}
\patchcmd{\ALG@doentity}{\item[]\nointerlineskip}{}{}{} 
\makeatother

\ifCLASSOPTIONcompsoc
  \usepackage[nocompress]{cite}
\else
  \usepackage{cite}
\fi

\ifCLASSINFOpdf
  \usepackage[pdftex]{graphicx}
  \graphicspath{{./}}
  \DeclareGraphicsExtensions{.pdf,.jpeg,.png}
\else
  \usepackage[dvips]{graphicx}
  \graphicspath{{./}}
  \DeclareGraphicsExtensions{.eps}
\fi

\usepackage{amsmath}
\ifCLASSOPTIONcompsoc
 \usepackage[caption=false,font=footnotesize,labelfont=sf,textfont=sf]{subfig}
\else
 \usepackage[caption=false,font=footnotesize]{subfig}
\fi

\begin{document}

\title{Enhancing Sound Source Localization\\
	via False Negative Elimination}

\author{Zengjie~Song,~\IEEEmembership{Member,~IEEE},
	    Jiangshe~Zhang,\\%
        Yuxi~Wang,
        Junsong~Fan,
        and~Zhaoxiang~Zhang,~\IEEEmembership{Senior~Member,~IEEE}
\IEEEcompsocitemizethanks{
\IEEEcompsocthanksitem Z. Song and J. Zhang are with the School of Mathematics and Statistics, Xi’an Jiaotong University, Xi’an 710049, China.\protect\\
Email: zjsong@hotmail.com, jszhang@mail.xjtu.edu.cn.%
\IEEEcompsocthanksitem Y. Wang and J. Fan are with the Center for Artificial Intelligence and Robotics, Hong Kong Institute of Science $\&$ Innovation, Chinese Academy of Sciences, Hong Kong, China. Email: yuxiwang93@gmail.com, junsong.fan@ia.ac.cn.%
\IEEEcompsocthanksitem Z. Zhang is with the New Laboratory of Pattern Recognition, State Key Laboratory of Multimodal Artificial Intelligence Systems, Institute of Automation, Chinese Academy of Sciences, Beijing 100190, China, also with the University of Chinese Academy of Sciences, Beijing 100049, China, and also with the Center for Artificial Intelligence and Robotics, Hong Kong Institute of Science $\&$ Innovation, Chinese Academy of Sciences, Hong Kong, China. Email: zhaoxiang.zhang@ia.ac.cn.
}
\thanks{Manuscript first version received August 20, 2022; resubmitted March 01, 2024; revised June 12, 2024; accepted August 08, 2024.}%
}

\markboth{Journal of \LaTeX\ Class Files,~Vol.~8, No.~20, August~2022}%
{Song \MakeLowercase{\textit{et al.}}: Enhancing Sound Source Localization}

\IEEEtitleabstractindextext{%
\begin{abstract}
	Sound source localization aims to localize objects emitting the sound in visual scenes. Recent works obtaining impressive results typically rely on contrastive learning. However, the common practice of randomly sampling negatives in prior arts can lead to the false negative issue, where the sounds semantically similar to visual instance are sampled as negatives and incorrectly pushed away from the visual anchor/query. As a result, this misalignment of audio and visual features could yield inferior performance. To address this issue, we propose a novel audio-visual learning framework which is instantiated with two individual learning schemes: self-supervised predictive learning (SSPL) and semantic-aware contrastive learning (SACL). SSPL explores image-audio positive pairs alone to discover semantically coherent similarities between audio and visual features, while a predictive coding module for feature alignment is introduced to facilitate the positive-only learning. In this regard SSPL acts as a negative-free method to eliminate false negatives. By contrast, SACL is designed to compact visual features and remove false negatives, providing reliable visual anchor and audio negatives for contrast. Different from SSPL, SACL releases the potential of audio-visual contrastive learning, offering an effective alternative to achieve the same goal. Comprehensive experiments demonstrate the superiority of our approach over the state-of-the-arts. Furthermore, we highlight the versatility of the learned representation by extending the approach to audio-visual event classification and object detection tasks. Code and models are available at: \href{https://github.com/zjsong/SACL}{https://github.com/zjsong/SACL}.
\end{abstract}

\begin{IEEEkeywords}
Sound source localization, audio-visual learning, contrastive learning, self-supervised learning, feature alignment
\end{IEEEkeywords}}

\maketitle

\IEEEraisesectionheading{\section{Introduction}\label{sec:introduction}}
\IEEEPARstart{W}{hen} strolling in a park brimming with life, you notice that the bird sitting on a twig is chirping; the puppy on the road ahead gives a little bark; and after a while an acquaintance may walk by and say hello to you friendly. Despite a short notice, humans own the excellent ability to associate the sounds they hear with the corresponding visual perception, and thus can localize and distinguish different sounding objects from one another. 

To mimic humans' such ability, in this work, we pay attention to the task of audio-visual sound source localization, where the goal is to localize regions of the visual landscape that correlate highly with the audio cues. While handling this task is a long-standing challenge \cite{Hershey99,Fisher00}, remarkable breakthroughs have been made until recent progresses on self-supervised audio-visual learning \cite{Owens18,Senocak18,Hu19,Qian20,Afouras20,Hu20b}. These methods leverage the free supervision rooted in videos, e.g., the natural correspondence and/or temporal synchronization between audio and visual sources, to guide multimodal feature extraction and alignment; then the similarity map between audio and visual features is usually employed to localize sounding objects. Among them, contrastive learning has particularly achieved impressive performance on this task \cite{Senocak18,Qian20,Afouras20,Chen21c,Tian21}.  

Existing contrastive learning methods in this line of work can be cast into two categories: the first one is global-level contrastive learning (GLCL) \cite{Senocak18,Senocak19,Morgado20,Qian20,Afouras20}, which commonly attracts audio and visual features extracted from the same video and repulses features from different videos; the other one is local-level contrastive learning (LLCL) \cite{Tian21,Chen21c,Lin21,Senocak22,Liu22}, which further compares audio feature with different visual feature components, even though they have correspondence at the video level. Generally, to perform contrastive learning, these methods \emph{randomly} sample sounds to form negative pairs with the given video frame. However, such randomness can produce false negatives by sampling sounds that actually belong to the same category with the positive sound, and thus hampering the model to align audio and visual features in semantic level. Consequently, the contrastive learning accompanied by feature misalignment would obtain inferior localization results.

\begin{figure*}
	\centering
	\includegraphics[width=\linewidth]{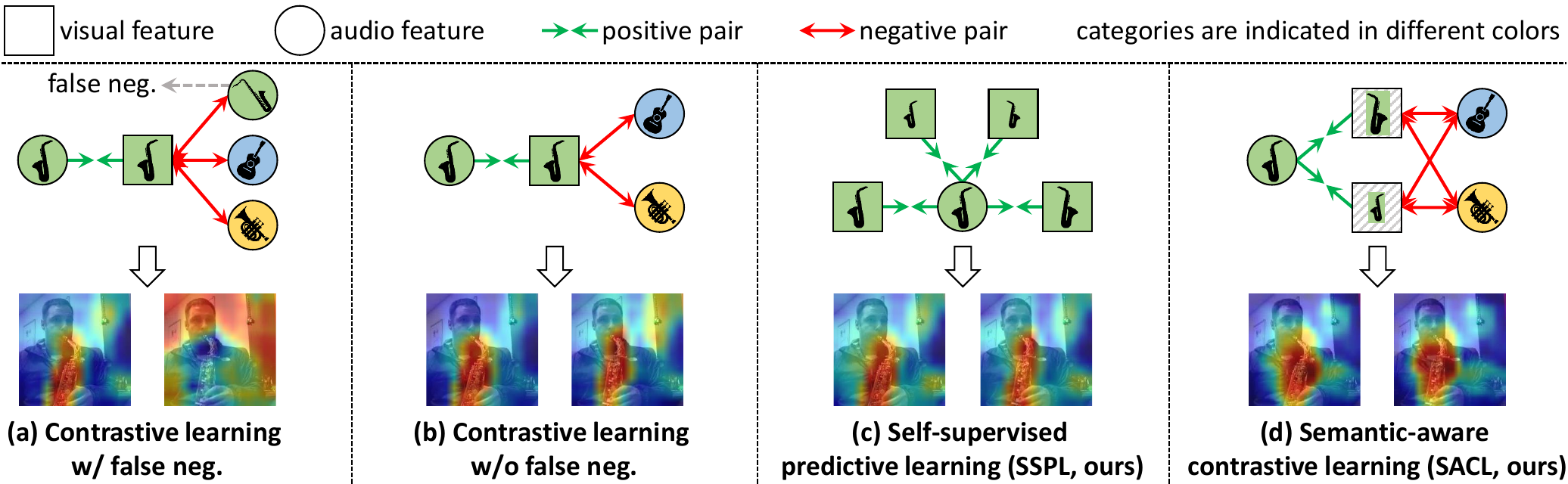}
	\caption{Effect of false negatives on sound localization. For two images that are visually similar but differ in local content (e.g., hand pose), \textbf{(a)} ambiguous localization is observed when performing contrastive learning \cite{Senocak18} with false negatives (i.e., audio samples that have the same class label with the positive). \textbf{(b)} Consistent localization can be obtained without sampling false negatives, but requiring class label as guidance. By contrast, our method mitigates the false negative problem via \textbf{(c)} self-supervised predictive learning (SSPL), which only relies on image-audio positive pair mining; or \textbf{(d)} semantic-aware contrastive learning (SACL), which proceeds with more reliable multimodal features by compacting visual features and meanwhile removing potential false negatives. Experiments are performed on MUSIC \cite{Zhao18}. See Section \ref{sec:ablation_sacl_false_neg} for quantitative comparisons.}%
	\label{fig:pilot_fasle_neg}
\end{figure*}
We carry out a pilot experiment to illustrate the effect of such false negatives in Fig. \ref{fig:pilot_fasle_neg}. Given image and sound from the same video (e.g., saxophone playing) as positive pair, other videos' sounds, holding the same category with the positive one, are allowed to construct negative pairs in Fig. \ref{fig:pilot_fasle_neg}a, while not allowed in Fig. \ref{fig:pilot_fasle_neg}b. We keep the remaining training settings same for these two cases. During test, consequently, the former case generates ambiguous localization on sounding objects (i.e., saxophone here) and the later one not. This raises the need to eliminate false negative samples, so as to release the full potential of learning-based sound localization. To this end, we shed light on two crucial questions: \emph{Can the image-audio positive pair alone be used to achieve this goal? Or, if we still shape sound localization within contrastive learning framework, can false negatives be effectively identified and removed via strategic negative sampling?}

In this paper, we answer these questions affirmatively by proposing a new audio-visual learning framework. The framework alleviates the false negative issue with two alternative learning schemes: self-supervised predictive learning (SSPL, sketch shown in Fig. \ref{fig:pilot_fasle_neg}c) and semantic-aware contrastive learning (SACL, sketch shown in Fig. \ref{fig:pilot_fasle_neg}d). Specifically, SSPL explicitly mines image-audio positive pairs by associating sound source with different augmented views of the given video frame, resulting in semantically coherent similarities between audio and visual features. In this regard, SSPL serves as a negative-free approach to implement sound localization, avoiding sampling false negatives indirectly. In addition, SACL performs contrastive learning on multimodal features, where the semantic visual features are compacted with pseudo object masks, while audio features that share similar semantic concepts with the positive sound are identified as false negatives and then removed during training. By doing so, SACL focuses on more faithful anchors/queries (i.e., compact visual features) and negatives (i.e., selected audio features), hence improving the efficacy of audio-visual contastive learning.

In summary, we make the following contributions:
\begin{itemize}
	\item We identify the false negative issue in contrastive learning-based sound localization through qualitative and quantitative experiments and analysis.
	
	\item We propose a unified framework that accommodates a negative-free method (SSPL, non-contrastive) and a negative-calibrated approach (SACL, contrastive) to mitigate the false negative issue.

	\item We devise a new contrastive learning scheme in audio-visual setting, effectively selecting compact visual anchor samples and reliable audio negative samples in a semantic-aware manner.
	
	\item We conduct extensive experiments to validate the superiority of our approach for sound localization, as well as its promising expandability on audio-visual event classification and object detection tasks.
\end{itemize}

This paper is an extension of our conference work in \cite{Song22}, where the basic idea of explicitly mining image-audio positive pairs for sound localization is introduced (i.e., SSPL, a \emph{non-contrastive} method). The present work differs from the conference version in four aspects. (1) We exploit a variant of the three-stream network used in SSPL to perform calibrated audio-visual \emph{contrastive} learning (i.e., SACL), which brings a new alternative way to competently mitigate the false negative problem. (2) The two approaches that learn representations with or without negatives are dialectically investigated and compared, offering a better understanding on how negatives influence the representation quality for sound localization. (3) We incorporate more state-of-the-art methods for methodology analysis and numerical comparisons, including two most recent works \cite{Morgado21b,Sun23} that also serve to alleviate the adverse influence of false negatives. (4) To show the generality and potential of the proposed learning schemes, we extend and evaluate our approach on another two typical applications (audio-visual event classification and self-supervised object detection).

\section{Related Work}\label{sec:related_work}
\subsection{Self-Supervised Visual Representation Learning}
Self-supervised learning (SSL) has achieved remarkable breakthroughs on large computer vision benchmarks. Many of the current SSL methods \cite{Henaff20,Chen20a,Chen20b,Chen20c,He20,Tian20,Feichtenhofer21,Chen21d,Qian21} resort to the design of contrastive learning rules \cite{Oord18}. These methods, at their core, transform one image into multiple views, and repulse different images (negatives) meanwhile attracting the same image’s different views (positives). While negative samples play an essential role in avoiding collapsed representations \cite{Henaff20,Chen20a,He20,Tian20}, they need to be maintained in a carefully designed memory bank \cite{Wu18,Tian20}, or in a queue generated from a momentum encoder \cite{He20,Chen20b}, or directly in a batch data of large batch size \cite{Chen20a,Chen20c}. Recently, several efforts have been made to further relieve the requirement of negatives and simplify the SSL framework beyond conventional contrastive learning, including BarlowTwins \cite{Zbontar21}, W-MSE \cite{Ermolov21}, BYOL \cite{Grill20}, and SimSiam \cite{Chen21a}. In SimSiam \cite{Chen21a}, researchers investigate the importance of simple Siamese architecture for unsupervised representation learning, and empirically show that the stop-gradient operation is critical for the network to prevent collapse, even without using momentum encoder \cite{He20} and large batches \cite{Chen20a}. These advances in image representation learning provide insights for our work to develop effective audio-visual representation learning method.

\subsection{Audio-Visual Representation Learning}
The vision and sound are usually two co-occurring modalities, which can naturally be used to derive supervisions for audio-visual learning \cite{Aytar16,Owens16,Korbar18,Chen20e,Su21}. In \cite{Aytar16}, for instance, the visual features extracted from pre-trained teacher networks act to guide the student network to learn more discriminative sound representation, and vice versa in \cite{Owens16}. Korbar et al. \cite{Korbar18} and Owens and Efros \cite{Owens18} leverage the synchronization between audio and visual streams to build negative samples and contrastive losses, obtaining versatile multisensory features, respectively. Several works also explore the audio-visual correspondence by feature clustering \cite{Hu19,Hu20a,Alwassel20}. In general, these methods focus on learning task-agnostic representations, which work well on classification-related down-stream tasks, such as action/scene recognition \cite{Korbar18,Owens18,Morgado20,Alwassel20,Morgado21a,Chen21b}, audio event classification \cite{Arandjelovic17,Korbar18,Hu19,Alwassel20,Morgado21b}, video retrieval \cite{Morgado21a,Chen21b}, \emph{etc}. However, they are not customized for sound source localization, and as a result only achieve limited performance on this task \cite{Owens18,Hu19,Hu20a}.

\subsection{Audio-Visual Sound Source Localization}
Early works to solve this task mainly rely on statistical modeling of the cross-modal relationship by using, for example, mutual information \cite{Hershey99,Fisher00} and canonical correlation analysis \cite{Kidron05,Izadinia12}. However, these approaches as shallow models only show advantages in simple audio-visual scenarios. By digging into the correspondence between deep audio and visual features, recent deep learning methods give promising solutions to this problem \cite{Senocak18,Hu19,Gan19,Qian20,Afouras20,Chen21c,Tian21,Valverde21,Hu21,Xuan22,Hu22}. For instance, Senocak et al. \cite{Senocak18} employ a two-stream framework and an attention mechanism to compute sound localization map. Qian et al. \cite{Qian20} achieve the same goal by using the class activation map derived from a weakly-supervised model. In \cite{Hu19,Hu20a}, audio and visual features are clustered, respectively, and the assignment weights based on the distance between features and cluster centers are adopted to localize sounding objects. In addition to viewing image and sound extracted from different videos as negative pair, Chen et al. \cite{Chen21c} and Lin et al. \cite{Lin21} propose to mine hard negatives within an image-audio pair, i.e., background regions exhibiting low correlation with the given sound are regarded as extra hard negatives. The contemporary work of \cite{Liu22} further extends the approach in \cite{Chen21c} with diverse audio and visual data augmentations. Different from these works that sample negatives randomly, our SSPL \cite{Song22} handles the same task by explicit positive mining, providing a novel and effective alternative for sound localization. What's more, our SACL investigates the strategy and vital influence of detecting and removing false negatives in audio-visual contrastive learning,  which is ignored in these work.

\begin{figure*}
	\centering
	\includegraphics[width=\textwidth]{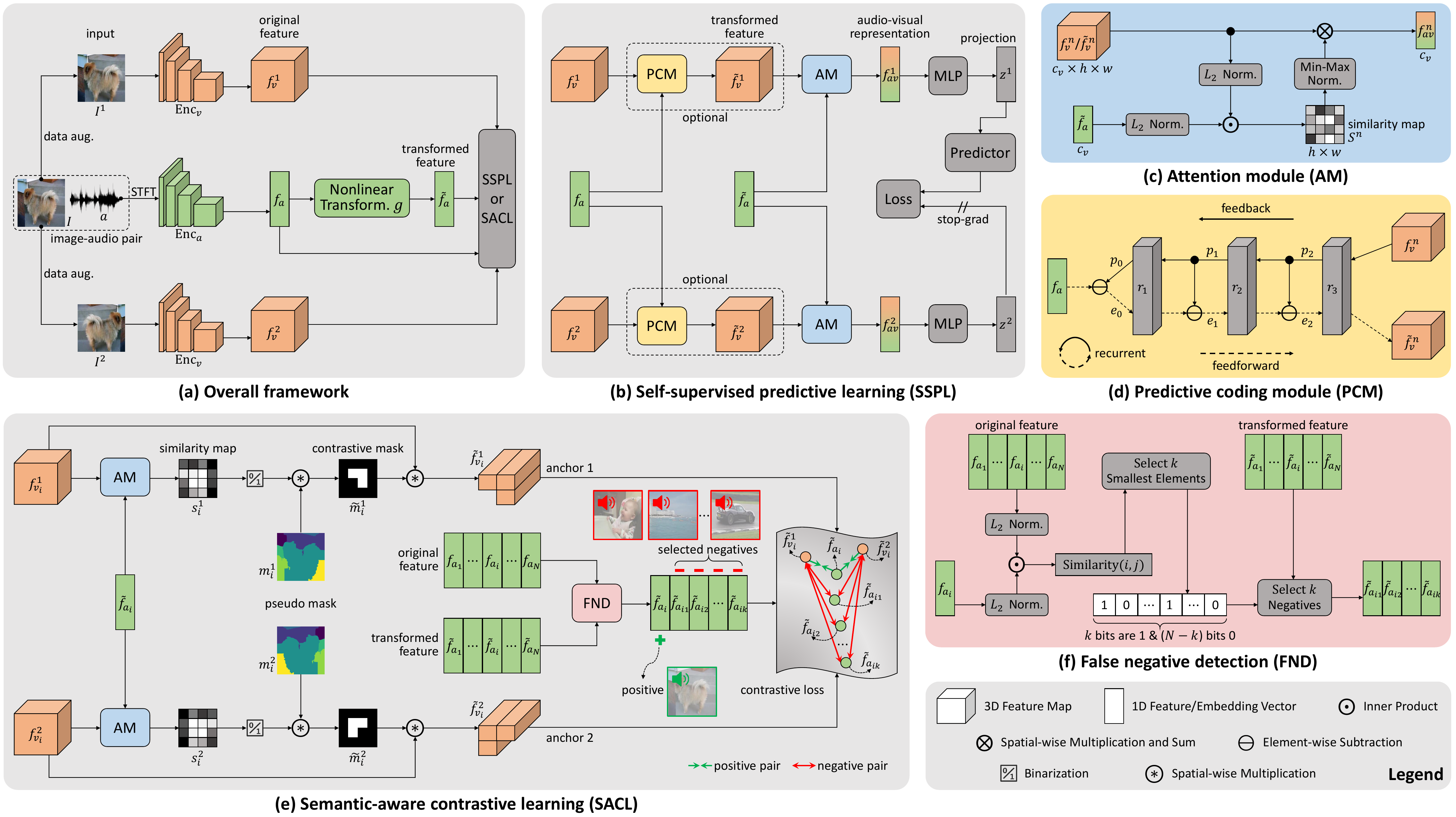}
	\caption{Illustration of our sound source localization method. The overall framework in \textbf{(a)} builds on a three-stream network: top and bottom streams for visual feature extraction and middle steam for audio signal processing. The audio-visual correspondence for sound localization is captured by two alternative learning schemes: self-supervised predictive learning (SSPL, Section \ref{sec:sspl}) and semantic-aware contrastive learning (SACL, Section \ref{sec:sacl}). SSPL in \textbf{(b)} is a negative-free approach, which aims to associate sound source with different augmented views of one image by mining image-audio pairs from the same video clip. To this end, the attention module (AM, Section \ref{sec:sspl_am}) in \textbf{(c)} for feature integration and the predictive coding module (PCM, Section \ref{sec:sspl_pcm}) in \textbf{(d)} for feature alignment are introduced. SACL in \textbf{(e)} is a contrastive learning method and works to find reliable anchor and negative features for contrast. The anchor point is determined by compacting visual features with pseudo mask and similarity map (Section \ref{sec:sacl_compact_vis_feat}), while the effective negatives are selected through the false negative detection (FND, Section \ref{sec:sacl_detect_fasle_neg}) module in \textbf{(f)}.}%
	\label{fig:method}
\end{figure*}

\subsection{False Negative Issue in Contrastive Learning}
Without access to labels, self-supervised contrastive learning randomly samples negative datapoints that, however, may be semantically related to the anchor point. These datapoints, termed as false negatives, are equally repelled from the anchor with all other negatives, which would adversely affect the representation learning and deteriorate downstream performance \cite{Arora19,Khosla20,Chuang20,Li21,Huynh22,Chen22}. To correct for the sampling of same-label datapoints, Chuang et al. \cite{Chuang20} develop a debiased contrastive objective by reweighing positive and negative terms, yet it does not identify false negatives. Huynh et al. \cite{Huynh22} propose to determine potential false negatives by comparing the similarities between each negative sample and all samples in a constructed support set, showing consistent improvements on ImageNet classification. Recently, Chen et al. \cite{Chen22} find that the performance drops induced by false negatives are more significant on large-scale datasets with more semantic contents. To handle this issue, they introduce a clustering-based contrastive learning framework that incrementally detects and eliminates false negative samples. Beyond these works that mainly concentrate on visual representation learning, there is, however, less progress made to tackle such issue in audio-visual setting. Particularly, based on the cross-modal and within-modal instance similarities, Morgado et al. \cite{Morgado21a,Morgado21b} design a soft instance discrimination loss to place less emphasis upon false negatives. Concurrent and independent works also find that removing/suppressing false negatives and taking a step forward by viewing them as hard positives \cite{Senocak22}, or by enhancing the impact of plausibly true negatives \cite{Sun23} in contrastive objective can improve sound localization. Nevertheless, the instance similarities computed in these methods are relatively unreliable in earlier training stage, and the use of false negatives identified by such fragile similarities would further disrupt the learning procedure. By contrast, our method (SACL) employs a pre-trained audio classification model to derive discriminative audio embedding, allowing false negative detection with sufficient confidence and thus leading to more reliable and stable contrastive learning.

\section{Methodology}\label{sec:method}
\subsection{Overview of Framework}\label{sec:overview}
Fig. \ref{fig:method}a depicts the overall framework of our sound localization approach, which features a three-stream network, making a big difference with widely-used two-stream ones. The top and bottom streams serve to extract deep visual features from different views of the same image, and the middle stream is responsible for deriving discriminative audio feature from the given sound source. After that, audio and visual features are leveraged to conduct audio-visual representation learning in two different and individual schemes, while the sound localization map is a natural consequence of representation learning and is generated in the attention module (AM).

Notably, the first learning scheme\textemdash self-supervised predictive learning (SSPL, Fig. \ref{fig:method}b)\textemdash builds audio-visual correspondence for localization by delving into image-audio pair from the same video clip, namely SSPL is a negative-free method. By contrast, the second learning scheme\textemdash semantic-aware contrastive learning (SACL, Fig. \ref{fig:method}e)\textemdash aims to find plausibly reliable anchor (visual modality) and negative (audio modality) samples, which is implemented by tailoring visual features with pseudo masks and removing false negatives based on audio feature similarities. In the following, we first introduce the general processes to extract original audio and visual features, and then elaborate on and formulate the two learning schemes, respectively.

\subsection{Unimodal Features of Audio and Vision}\label{sec:unimodaal_feature}
Let $I \in \mathbb{R}^{3\times H_{v}\times W_{v}}$ and $a\in\mathbb{R}^{H_{a}\times W_{a}}$ denote a video frame and a corresponding audio signal from the same video clip, respectively. Here the raw 1D audio waveform has been converted into the 2D spectrogram by Short-Time Fourier Transform (STFT), and therefore we use 2D CNNs to extract deep semantic features of audio modality like vision. In practice, we employ the off-the-shelf VGG16 \cite{Simonyan15} for frame processing ($\text{Enc}_{v}$) in SSPL, and ResNet-18 \cite{He16} in SACL, while utilizing the VGGish network \cite{Hershey17} for spectrogram analysis ($\text{Enc}_{a}$), similar to \cite{Hu20a}. The output feature map of the final convolution layer of VGG16/ResNet-18 is treated as the original visual feature $f_{v}$. We use layers before the final post-processing stage of VGGish to produce a high-level embedding as the original audio feature $f_{a}$. These feature extraction processes are formulated as:
\begin{align}
	f_{v} &= \text{Enc}_{v}(I), \quad f_{v}\in \mathbb{R}^{c_{v}\times h\times w}, \label{eq:orig_visual_feat}\\
	f_{a} &= \text{Enc}_{a}(a), \quad f_{a}\in \mathbb{R}^{c_{a}}. \label{eq:orig_audio_feat}
\end{align}

Let $I^{1}$ and $I^{2}$ denote two randomly augmented views of the given image $I$. The two views are respectively fed into the visual CNN, $\text{Enc}_{v}$, to obtain spatial feature maps $f_{v}^{1}$ and $f_{v}^{2}$ as in \eqref{eq:orig_visual_feat}. For simplicity, we use $n\in\{1,2\}$ to index different visual views. Considering that $f_{a}$ and $f^{n}_{v}$ are from two heterogeneous modalities, we further transform $f_{a}$ to be comparable with the visual feature via a nonlinear transformation $g$, i.e., $\tilde{f}_{a}=g(f_{a})\in\mathbb{R}^{c_{v}}$. Equipped with unimodal features, we are able to perform consequent representation learning in two different ways.

\subsection{Self-Supervised Predictive Learning}\label{sec:sspl}
The first learning scheme, SSPL, is a negative-free method for sound localization. Here, our hypothesis is that two visual scenes containing the same sounding objects should consistently correspond to the same audio cues in semantic level. To this end, as shown in Fig. \ref{fig:method}b, SSPL integrates audio and visual features to form the audio-visual representation through an attention module (AM, Fig. \ref{fig:method}c), and then the two audio-visual representations are enforced to be similar by self-supervised predicting with each other. We also design the predictive coding module (PCM, Fig. \ref{fig:method}d) for audio-visual feature alignment. It is noteworthy that the vanilla SSPL without PCM focuses on exploring audio-visual correspondence across different image views (Sections \ref{sec:sspl_am} and \ref{sec:sspl_loss}), while the PCM component excels at aligning features across modalities (Section \ref{sec:sspl_pcm}), and thus boosting localization performance further.

\subsubsection{Attention Module}\label{sec:sspl_am}
We adopt the $L_{2}$-normalized inner product (or cosine similarity) to measure the similarity between audio and visual features, as suggested by \cite{Senocak18,Chen21c}. Formally, for the spatial location $(i,j)$ in visual feature map $f_{v}^{n}$, a similarity value is computed as follows:
\begin{equation}\label{eq:cosine_similarity}
	S^{n}(i,j) = \frac{\langle \tilde{f}_{a}, f_{v}^{n}(\cdot,i,j) \rangle}{\|\tilde{f}_{a}\|_{2}\|f_{v}^{n}(\cdot,i,j)\|_{2}}, \quad (i,j) \in [h]\times[w],
\end{equation}
where $f_{v}^{n}(\cdot,i,j)\in \mathbb{R}^{c_{v}}$. 

The similarity map $S\in\mathbb{R}^{h\times w}$ plays two important roles in our method. On the one hand, it indicates the degree of correlation between each image location (after resized to image scale) and the given audio cues, and thus can serve as the sound localization map. On the other hand, it acts as an attention mechanism to weigh the original visual feature, resulting in the following audio-visual representation:
\begin{equation}\label{eq:av_repres}
	f_{av}^{n}(k) = \sum_{i,j}\tilde{S}^{n}(i,j)f_{v}^{n}(k,i,j), \quad k\in\{1,\dots,c_{v}\},
\end{equation}
where
\begin{equation}\label{eq:norm_sim_map}
	\tilde{S}^{n} = \frac{S^{n}-\min(S^{n})}{\max(S^{n})-\min(S^{n})}.
\end{equation}
Here we scale the similarity values to $[0,1]$ by min-max normalization \cite{Lin21}. This operation makes different feature elements more distinguishable, and performs better compared with the sigmoid and softmax scaling functions \cite{Senocak18,Qian20} (see Table \ref{tab:ablation_sspl_scaling} for empirical comparisons). Since the $f_{av}^{n}\in \mathbb{R}^{c_{v}}$ selects and integrates visual features that are more related to audio ones, we treat it as a multimodal representation to advance subsequent learning.

\subsubsection{SSPL Loss}\label{sec:sspl_loss}
The learning procedure aims to make the two audio-visual representations similar. We follow SimSiam \cite{Chen21a} to achieve this goal in the audio-visual setting. Formally, we feed $f_{av}^{n}$ into a MLP head to obtain the projection of corresponding view, $z^{n}=\text{MLP}(f_{av}^{n})$. Then a predictor head, denoted as Pred, takes as input $z^{1}$ to predict $z^{2}$ by minimizing the negative cosine similarity (NCS):
\begin{equation}\label{eq:neg_cos_sim}
	\mathcal{L}_{NCS}(z^{1}, z^{2}) = -\frac{\langle \text{Pred}(z^{1}), z^{2} \rangle}{\|\text{Pred}(z^{1})\|_{2}\|z^{2}\|_{2}}.
\end{equation}
To symmetrize the above loss, we also feed $z^{2}$ into Pred to estimate $z^{1}$, leading to another loss term $\mathcal{L}_{NCS}(z^{2}, z^{1})$. The total loss is therefore defined as:
\begin{equation}\label{eq:simsiam_loss0}
	\mathcal{L}_{SSPL} = \frac{1}{2}\mathcal{L}_{NCS}(z^{1}, z^{2}) + \frac{1}{2}\mathcal{L}_{NCS}(z^{2}, z^{1}).
\end{equation} 
However, as discussed in \cite{Chen21a}, directly minimizing the loss in Eq. \eqref{eq:simsiam_loss0} could easily induce representation collapse. To overcome this problem the stop-gradient (SG) operation is employed. That is, Eq. \eqref{eq:neg_cos_sim} is modified as $\mathcal{L}_{NCS}(z^{1}, \text{SG}(z^{2}))$, where $z^{2}$ is viewed as a constant such that branch on $I^{2}$ receives no gradient from $z^{2}$ through this loss term. Similarly we have $\mathcal{L}_{NCS}(z^{2}, \text{SG}(z^{1}))$, and the form in Eq. \eqref{eq:simsiam_loss0} is implemented as:
\begin{equation}\label{eq:simsiam_loss}
	\mathcal{L}_{SSPL} = \frac{1}{2}\mathcal{L}_{NCS}(z^{1}, \text{SG}(z^{2})) + \frac{1}{2}\mathcal{L}_{NCS}(z^{2}, \text{SG}(z^{1})).
\end{equation}

\subsubsection{Predictive Coding Module}\label{sec:sspl_pcm}
In this section, we propose the PCM for audio and visual feature alignment, and continuously improving the localization performance of SSPL. The key idea inherits the spirit of predictive coding (PC) in neuroscience \cite{Rao99,Spratling17,Wen18,Song18,Song23}, which simulates the mechanism of information processing in visual cortex. In brief, PC uses feedback connections from a higher-level area to a lower-level one to convey predictions of lower-level neural activities; it employs feedforward connections to carry the errors between the actual activities and the predictions; and the brain dynamically updates representations so as to progressively reduce the prediction errors. In our PCM (Fig. \ref{fig:method}d), we treat the visual feature as a type of prior knowledge to predict the audio feature in an \emph{iterative} manner. In the following, at the heart of PCM, we give the representation update rules of feedback and forward processes, respectively. The detailed derivations can be found in the supplement.

Denote by $r_{l}(t), l\in\{1,\dots,L\}, t\in\{0,\dots,T\}$ the representation of the $l$-th layer of PCM network at time step $t$, and by $W_{l,l-1}$ the feedback connection weights from layer $l$ to layer $l-1$ (and vice versa for $W_{l-1,l}$).

The \emph{feedback process} updates representations through a mechanism of layer-wise prediction generation. Concretely, at $l$-th layer the prediction, $p_{l}$, of representation, $r_{l}$, is first derived using the above layer's representation, $r_{l+1}$. Then $r_{l}$ is updated with its previous state and the prediction, i.e., at time step $t$ we have:
\begin{align}
	p_{l}(t)   &= (W_{l+1,l})^{T}r_{l+1}(t), \label{eq:feedback_pred}\\
	r_{l}(t) &\gets \phi((1-b_{l})r_{l}(t-1) + b_{l}p_{l}(t)), \label{eq:feedback_repres} 
\end{align}
where $\phi$ is a nonlinear activation function and $b_{l}$ serves as a positive scalar to balance two terms. The above update rules are executed from top layer $L$ to bottom layer $1$ in sequence, and by setting $p_{L}(t)\equiv f_{v}^{n}$, we in fact achieve further feature extraction from visual source. 

In \emph{feedforward process}, representations are again modulated based on prediction errors emerged at each layer. Specifically, the representation $r_{l-1}$ and its prediction $p_{l-1}$ are often unequal, resulting in a prediction error $e_{l-1}$. The error signal contains unpredictable components of $r_{l-1}$, and is forwarded to higher level to correct the representation $r_{l}$. This leads to complementary update rules:
\begin{align}
	e_{l-1}(t) &= r_{l-1}(t) - p_{l-1}(t), \label{eq:feedforward_error}\\
	r_{l}(t) &\gets \phi(r_{l}(t) + a_{l}(W_{l-1,l})^{T}e_{l-1}(t)), \label{eq:feedforward_repres}
\end{align}
where $r_{0}(t)\equiv f_{a}$ is the original audio feature, $p_{0}(t)=\phi((W_{1,0})^{T}r_{1}(t))$ refers to the prediction of $f_{a}$, and $a_{l}$ denotes a trade-off scalar like $b_{l}$.

PCM conducts the two distinct processes alternatively while all layers' representations are progressively refined so as to reduce the prediction error. Subsequently, we use a $1\times1$ convolution to transform the top layer representation at last time step, $r_{L}(T)$, to a new visual feature, $\tilde{f}_{v}^{n}$, with the same dimension of $f_{v}^{n}$. We summarize the main computing steps in Algorithm \ref{alg:update_repres}. As such, the vanilla SSPL method can be enhanced by feeding $\tilde{f}_{v}^{n}$, instead of $f_{v}^{n}$, into the AM to compute audio-visual representation (i.e., Eqs. \eqref{eq:cosine_similarity} and \eqref{eq:av_repres}).
\begin{algorithm}[t]
	\caption{Update Representations in PCM}\label{alg:update_repres}
	\begin{algorithmic}[1]
		\renewcommand{\algorithmicrequire}{\textbf{Input:}}
		\renewcommand{\algorithmicensure}{\textbf{Output:}}
		\Require $f_{v}^{n}$ and $f_{a}$
		\Ensure $\tilde{f}_{v}^{n}$
		\vspace{1mm}
		\For {$t=0$ to $T$}
		\If {$t=0$}
		\State initialize representations
		\EndIf
		
		\For {$l=L$ to $1$}\Comment{feedback process}
		\If {$l=L$}
		\State $p_{l}(t) = f_{v}^{n}$
		\Else
		\State compute prediction $p_{l}(t)$: Eq. \eqref{eq:feedback_pred}
		\EndIf
		\State update representation $r_{l}(t)$: Eq. \eqref{eq:feedback_repres}
		\EndFor
		
		\For {$l=1$ to $L$}\Comment{feedforward process}
		\If {$l=1$}
		\State $e_{l-1}(t) = f_{a} - \phi((W_{l,l-1})^{T}r_{l}(t))$
		\Else
		\State obtain prediction error $e_{l-1}(t)$: Eq. \eqref{eq:feedforward_error}
		\EndIf
		\State update representation $r_{l}(t)$: Eq. \eqref{eq:feedforward_repres}
		\EndFor
		\EndFor
		
		\State $\tilde{f}_{v}^{n}=\texttt{Conv}_{1\times1}(r_{L}(T))$\Comment{transformed feature}
	\end{algorithmic}
\end{algorithm}

\subsection{Semantic-Aware Contrastive Learning}\label{sec:sacl}
With explicit positive mining, SSPL indirectly avoids the false negative problem induced by random negative sampling. In this section, by contrast, we present another alternative learning scheme, SACL, to directly and effectively tackle such unfavorable problem. The core idea underlying SACL is to find reliable visual and audio features as anchor and negative points, respectively, and thus enabling the contrastive learning to build more accurate cross-modal correspondence in semantic level. As exhibited in Fig. \ref{fig:method}e, on the one hand, SACL leverages the intersection of similarity map and pseudo mask to extract visual features that are semantically compact. On the other hand, SACL takes advantage of the audio feature similarity to detect potential false negatives, and then constructs the reliable negative set by removing these false negatives from current mini-batch (Fig. \ref{fig:method}f).

\subsubsection{Visual Feature Compaction}\label{sec:sacl_compact_vis_feat}
\begin{figure}
	\centering
	\includegraphics[width=\linewidth]{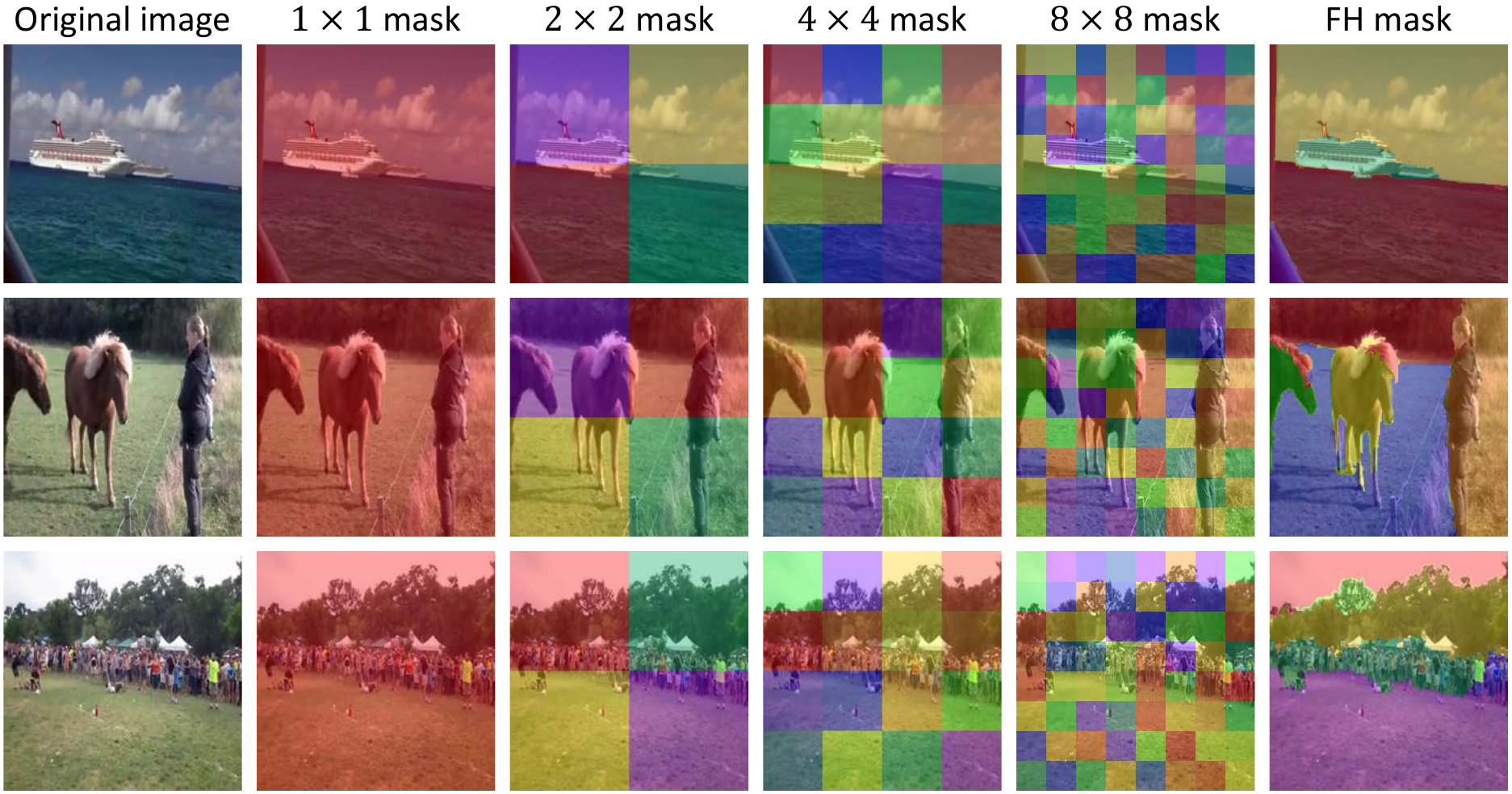}%
	\caption{Example pseudo masks used in SACL. In addition to the image-computable FH masks (6th column), we also consider spatial heuristics that partition the image into $2\times2$, $4\times4$, $8\times8$ grids (3rd-5th columns, respectively). Note that a $1\times1$ grid (2nd column) is equivalent to only using the binary similarity map to compact visual features.}
	\label{fig:pseudo_mask_eg}
\end{figure}

In conventional audio-visual contrastive learning, visual features of the \emph{whole} image are usually contrasted with the audio ones. Due to the presence of background distractions, such global-level contrast could hinder the learning process from finding fine-grained audio-visual correlation. Here, inspired by \cite{Henaff2021}, we introduce a simple mask-guided approach to identify object-based regions, by which tailored visual features can be more compact and less ambiguous.

Specifically, SACL first employs an approximate, image-computable segmentation algorithm \cite{Felzenszwalb04} to generate coarse segmentation map of the raw image. The segmentation map is further augmented with the same spatial augmentations operated on the image counterpart, resulting in two pseudo masks $m^{1}$ and $m^{2}$. After that, the intersection of the binary similarity map (binarized by comparing each value with the similarity median), $\text{Bi}(S^{n})$, and pseudo mask, $m^{n}$, is computed as the contrastive mask:
\begin{align}
	\tilde{m}^{n} &= \text{Bi}(S^{n})\circledast m^{n}(j^{*}), \label{eq:contrast_mask}\\
	j^{*} &= \operatorname*{argmax}_{j} \text{Sum}(\text{Bi}(S^{n})\circledast m^{n}(j)),
\end{align}
where $\circledast$ denotes element- or spatial-wise multiplication, and $m^{n}(j)$ is the $j$-th binary sub-mask in $m^{n}$, such that $m^{n} = \bigcup_{j} m^{n}(j)$ and $m^{n}(p)\bigcap m^{n}(q)=\emptyset, \forall p\neq q$. We then use the contrastive mask to extract compacted visual feature:
\begin{equation}\label{eq:compact_vis_feat}
	\tilde{f}_{v}^{n} = \tilde{m}^{n}\circledast f_{v}^{n}.
\end{equation}
Intuitively, the contrastive mask identifies the region that has the biggest area of intersection of similarity map and pseudo mask. Hence, it is the visual feature in this region that carries both the high correlation with sound source (from similarity map) and the semantics of sounding object (from pseudo mask). We view the compact visual feature $\tilde{f}_{v}^{n}$ as the anchor point for contrastive learning.

To generate pseudo masks for visual feature compaction, we investigate several segmentation strategies, as shown in Fig. \ref{fig:pseudo_mask_eg}, from spatial heuristics to the unsupervised Felzenszwalb-Huttenlocher (FH) algorithm \cite{Felzenszwalb04}, like \cite{Henaff2021}. Notice that the FH algorithm has already been embedded in \texttt{scikit-image} \cite{van14}, therefore FH masks can be effortlessly generated during training. We perform ablation on the types of pseudo masks in Section \ref{sec:ablation_sacl_mask}.

\subsubsection{False Negative Detection}\label{sec:sacl_detect_fasle_neg}
For random negative sampling, audio features of those sounds, which are extracted from other videos different from the current video, are all allowed to construct negative pairs with the visual anchor feature. As a downside, such random sampling sometimes yields false negatives that hold the same category as the positive one, making contrastive learning undesirably repel the anchor from semantically similar audio samples. To address this issue, SACL directly detects and explicitly removes the false negatives according to the audio feature similarity within a mini-batch, as illustrated in Fig. \ref{fig:method}f.

Formally, given a batch of the original audio features $\{f_{a_{j}}\}_{j=1}^{N}$ and transformed audio features $\{\tilde{f}_{a_{j}}\}_{j=1}^{N}$, we first compute the $L_{2}$-normalized inner product of $f_{a_{i}}$ and $f_{a_{j}}$, obtaining similarity:
\begin{equation}\label{eq:audio_sim}
	S_{aa}(i,j) = \frac{\langle f_{a_{i}}, f_{a_{j}} \rangle}{\|f_{a_{i}}\|_{2}\|f_{a_{j}}\|_{2}}, \quad j = 1, \cdots, N.
\end{equation}
Because the original audio feature is actually the audio embedding derived from the pre-trained audio classification model \cite{Hershey17}, $S_{aa}(i,j)$ can approximate the true similarity between audio samples. We then conform the top-$k$ smallest elements in $\{S_{aa}(i,j)\}_{j=1}^{N}$, forming a binary index mask, where 1 represents a semantically dissimilar audio sample and 0 a similar sample (i.e., viewed as false negative). At last, the $k$ transformed audio features filtered by the binary index mask are grouped as final negatives $\{\tilde{f}_{a_{ij}}\}_{j=1}^{k}$, relative to the positive $\tilde{f}_{a_{i}}$. By doing so, we identify the remaining $N-k-1$ negatives as false negative candidates and eliminate them during contrastive learning.

\subsubsection{SACL Loss}\label{sec:sacl_loss}
We modulate the InfoNCE \cite{Oord18} to implement audio-visual contrastive learning. To further reduce the influence of background distractions, we use the (negative) maximum cosine similarity between transformed visual and audio features to quantify the cross-modal feature distance:
\begin{equation}\label{eq:max_sim}
	S_{va}^{n}(i,j) = \max_{u,v}\frac{\langle \tilde{f}_{v_{i}}^{n}(u,v), \tilde{f}_{a_{j}} \rangle}{\|\tilde{f}_{v_{i}}^{n}(u,v)\|_{2}\|\tilde{f}_{a_{j}}\|_{2}},
\end{equation}   
where $\tilde{f}_{v_{i}}^{n}(u,v)\in \mathbb{R}^{c_{v}}$. In final contrastive loss, the distance between visual and audio features extracted from the same video, $-S_{va}^{n}(i,i)$, is minimized, while $-S_{va}^{n}(i,ij)$ is maximized given that the $ij$-th audio feature, $\tilde{f}_{a_{ij}}$, belongs to the constructed negative set. These make for the following SACL loss:
\begin{equation}\label{eq:sacl_loss}
	\mathcal{L}_{SACL} = \frac{1}{N}\sum_{i=1}^{N}(\ell_{i}^{1} + \ell_{i}^{2}),
\end{equation}
where
\begin{equation}\label{eq:sacl_single_view_loss}
	\ell_{i}^{n} = -\log\frac{\exp(S_{va}^{n}(i,i)/\tau)}{\exp(S_{va}^{n}(i,i)/\tau) + \sum_{j=1}^{k}\exp(S_{va}^{n}(i,ij)/\tau)},
\end{equation}
where $\tau$ is the temperature hyper-parameter.

\section{Experiments}\label{sec:experiments}
\subsection{Datasets and Evaluation Protocol}\label{sec:dataset_eval}
\textbf{SoundNet-Flickr \cite{Aytar16}.} This dataset consists of more than 2 million videos from Flickr. We use a 3s audio clip around the middle frame of the whole audio, and the accompanied video frame to form an image-audio pair. Following \cite{Senocak18,Chen21c}, we train models with two random subsets of 10k and 144k image-audio pairs, respectively, and perform evaluation on the 250 annotated pairs provided by \cite{Senocak18}. Note that the location of the sound source in each test frame is given by 3 separate bounding boxes, each of which is obtained by a different annotator.

\textbf{VGG-Sound \cite{Chen20d} and VGG-Sound Source \cite{Chen21c}.} VGG-Sound dataset contains over 200k video clips that are divided into 300 sound categories. Similar to \cite{Chen21c}, we conduct training with 10k and 144k image-audio pairs randomly sampled from this dataset, respectively. For fair comparisons with recent works \cite{Senocak18,Afouras20,Chen21c,Senocak22,Liu22}, we evaluate models on the VGG-Sound Source (VGG-SS) benchmark with 5k annotated image-audio pairs collected by \cite{Chen21c}. Compared with SoundNet-Flickr benchmark that spans about 50 sounding object classes, VGG-SS has 220 classes and thus providing a more challenging scenario for sound localization task.

\textbf{Compared Methods.} We compare our approach with several typical sound localization methods, including Attention \cite{Senocak18}, DMC \cite{Hu19}, MSSL \cite{Qian20}, DSOL \cite{Hu20b}, HardWay \cite{Chen21c}, AVID \cite{Morgado21b}, HardPos \cite{Senocak22}, TIE \cite{Liu22}, and FNAC \cite{Sun23}. In particular, we have reproduced four related methods. The first two are Attention and HardWay, which could be representatives of GLCL- and LLCL-based approaches, respectively, and both of them sample negatives randomly for contrast. The other two are AVID and FNAC, which also focus on the false negative issue like ours. We denote our negative-free approach without using PCM by SSPL (w/o PCM), and the version equipped with PCM by SSPL (w/ PCM).

\textbf{Metrics.} We employ consensus Intersection over Union (cIoU) and Area Under Curve (AUC) as evaluation metrics, and report cIoU scores with threshold 0.5 in experiments, same as \cite{Senocak18,Qian20,Chen21c,Lin21,Senocak22,Liu22}. For both metrics, the higher value indicates more precise localization results.

\begin{table}
	\tabcolsep=8.8pt
	\centering
	\caption{Quantitative Results of Sound Localization on SoundNet-Flickr}%
	\label{tab:compare_flickr}
	\begin{threeparttable}		
		\begin{tabularx}{\linewidth}{@{}llcc}
			\toprule
			Method    							    				&Pre-train  			&cIoU $\uparrow$       			&AUC $\uparrow$ \\
			\midrule\midrule
			\multicolumn{4}{l}{\emph{Training on Flickr10k}}\\
			Attention \cite{Senocak18}$_{\text{CVPR}18}^{*}$  		&Yes, InitParam+Signal  &0.442                 			&0.461 \\
			DMC \cite{Hu19}$_{\text{CVPR}19}$						&Yes, InitParam+Signal  &0.414				   			&0.450 \\
			CAVL \cite{Hu20a}$_{\text{arXiv}20}$					&No	          			&0.500				   			&0.492 \\
			MSSL \cite{Qian20}$_{\text{ECCV}20}$      			 	&Yes, InitParam+Signal  &0.522    			   			&0.496 \\
			AVObject \cite{Afouras20}$_{\text{ECCV}20}$  			&No	          			&0.546 			       			&0.504 \\
			DSOL \cite{Hu20b}$_{\text{NeurIPS}20}$					&Yes, InitParam			&0.566 				   			&0.515 \\
			HardWay \cite{Chen21c}$_{\text{CVPR}21}^{*}$   			&Yes, InitParam			&0.615 				   			&0.535 \\
			AVID \cite{Morgado21b}$_{\text{CVPR}21}^{*}$            &No	          			&0.667							&0.552 \\
			ICL \cite{Lin21}$_{\text{CVPRW}21}$					 	&Yes, InitParam			&0.710				   			&0.580 \\
			TIE \cite{Liu22}$_{\text{ACM MM}22}$					&No	          			&\underline{0.755}              &0.588 \\
			FNAC \cite{Sun23}$_{\text{CVPR}23}^{*}$					&Yes, InitParam+Signal  &0.751				   			&\underline{0.604} \\
			SSPL (w/o PCM)   					 					&Yes, InitParam			&0.671				   			&0.556 \\
			SSPL (w/ PCM)    					 					&Yes, InitParam			&0.743                 			&0.587 \\
			SACL             					 					&Yes, InitParam+Signal  &\textbf{0.815}    	            &\textbf{0.623} \\
			\midrule
			\multicolumn{4}{l}{\emph{Training on Flickr144k}}\\
			Attention \cite{Senocak18}$_{\text{CVPR}18}$  		 	&Yes, InitParam+Signal  &0.660	   			   			&0.558 \\
			DMC \cite{Hu19}$_{\text{CVPR}19}$						&Yes, InitParam+Signal  &0.671 				   			&0.568 \\
			HardWay \cite{Chen21c}$_{\text{CVPR}21}^{*}$	 		&Yes, InitParam			&0.699				   			&0.590 \\
			AVID \cite{Morgado21b}$_{\text{CVPR}21}^{*}$            &No	          			&0.695							&0.563 \\
			HardPos \cite{Senocak22}$_{\text{ICASSP22}}$			&Yes, InitParam+Signal  &0.752                 			&0.597 \\
			TIE \cite{Liu22}$_{\text{ACM MM}22}$					&No	          			&\underline{0.815}   	        &0.611 \\
			FNAC \cite{Sun23}$_{\text{CVPR}23}^{*}$					&Yes, InitParam+Signal  &0.763				   			&\underline{0.618} \\
			SSPL (w/o PCM)   					 					&Yes, InitParam			&0.699				   			&0.580 \\
			SSPL (w/ PCM)    					 					&Yes, InitParam			&0.759  			   			&0.610 \\
			SACL             					 					&Yes, InitParam+Signal  &\textbf{0.851}    	            &\textbf{0.623} \\
			\midrule
			\multicolumn{4}{l}{\emph{Training on VGG-Sound10k}}\\
			Attention \cite{Senocak18}$_{\text{CVPR}18}^{*}$	 	&Yes, InitParam+Signal  &0.522                 			&0.502 \\
			HardWay \cite{Chen21c}$_{\text{CVPR}21}^{*}$   			&Yes, InitParam			&0.647                 			&0.560 \\
			AVID \cite{Morgado21b}$_{\text{CVPR}21}^{*}$            &No	          			&0.703							&0.559 \\
			FNAC \cite{Sun23}$_{\text{CVPR}23}^{*}$					&Yes, InitParam+Signal  &\underline{0.791}	   			&\underline{0.627} \\
			SSPL (w/o PCM)   					 					&Yes, InitParam			&0.699  			   			&0.572 \\
			SSPL (w/ PCM)    					 					&Yes, InitParam			&0.763						   	&0.591 \\
			SACL             					 					&Yes, InitParam+Signal  &\textbf{0.839}    				&\textbf{0.631} \\
			\midrule
			\multicolumn{4}{l}{\emph{Training on VGG-Sound144k}}\\
			HardWay \cite{Chen21c}$_{\text{CVPR}21}^{*}$	 		&Yes, InitParam			&0.723	   			   			&0.605 \\
			AVID \cite{Morgado21b}$_{\text{CVPR}21}^{*}$            &No	          			&0.771							&0.590 \\
			HardPos \cite{Senocak22}$_{\text{ICASSP22}}$			&Yes, InitParam+Signal  &0.768  			   			&0.592 \\
			TIE \cite{Liu22}$_{\text{ACM MM}22}$					&No	          			&0.795						   	&0.612 \\
			FNAC \cite{Sun23}$_{\text{CVPR}23}^{*}$					&Yes, InitParam+Signal  &\underline{0.819}				&\underline{0.627} \\
			SSPL (w/o PCM)   					 					&Yes, InitParam			&0.739	               			&0.602 \\
			SSPL (w/ PCM)    					 					&Yes, InitParam			&0.767                 			&0.605 \\
			SACL             					 					&Yes, InitParam+Signal  &\textbf{0.864}    				&\textbf{0.655} \\
			\bottomrule
		\end{tabularx}%
		\begin{tablenotes}[para,flushleft]
			\emph{``Pre-train'': whether using supervised pre-trained models; ``Yes, InitParam'': the pre-trained models serve to initialize parameters; ``Yes, InitParam+Signal'': the pre-trained models provide not only parameter initialization but also supervision signals; ``No'': training model in a fully self-supervised learning way. ``$*$'' denotes our reproduction. The best and runner-up results are marked in \textbf{bold} and \underline{underline}, respectively. These notes are the same for Tables \ref{tab:compare_vggss}-\ref{tab:av_object_detect}. }
		\end{tablenotes}%
	\end{threeparttable}
\end{table}

\subsection{Implementation Details}\label{sec:implement_detail}
\textbf{Common Setup.} The VGG16 \cite{Simonyan15} and ResNet-18 \cite{He16} pre-trained on ImageNet \cite{Deng09} are used as the backbone to extract semantic visual features in SSPL and SACL, respectively. The visual input is an image of size $256\times256$. We use VGGish \cite{Hershey17} pre-trained on AudioSet \cite{Gemmeke17} as the audio feature extractor. The raw 3s audio signal is re-sampled at 16kHz and further transformed into $96\times64$ log-mel spectrograms as the audio input, and the audio output $f_{a}$ is a 128D feature vector. The nonlinear audio feature transformation function $g(\cdot)$ is instantiated with a simple two-layer network as in \cite{Senocak18}: \texttt{FC}(512)-\texttt{ReLU}-\texttt{FC}(512). Additionally, we optimize the model with AdamW \cite{Loshchilov19}, while utilizing the early stopping strategy to avoid overfitting in all cases.

\textbf{SSPL.} The simple data augmentation pipeline is performed on input image: random cropping with $224\times224$ resizing and random horizontal flip. We closely follow SimSiam \cite{Chen21a} to set the projection and prediction MLPs. For PCM, we mainly adopt \texttt{Conv}-\texttt{MaxPool}-\texttt{GELU} layers in the feedback pathway, and \texttt{Upsample}-\texttt{DeConv}-\texttt{GELU} layers in the feedforward counterpart. The weights of two feature extractors are kept frozen during training.

\textbf{SACL.} The same set of image augmentations in SimCLR \cite{Chen20a} and SimSiam \cite{Chen21a} is used in our contrastive learning scenario, namely the two spatial augmentations in SSPL are followed by the color distortion, the grayscale conversion, and the Gaussian blur, all of which are applied randomly to the images. Note that only the two spatial augmentations are also operated on pseudo masks for semantic consistency. We generate the FH masks on the fly by fixing three hyper-parameters: scale $s=1000$, width of Gaussian kernel $\sigma=0.5$, and minimum cluster size $c=1000$. The temperature in SACL loss is also fixed in all experiments, i.e., $\tau=0.005$. What's more, SACL allows parameter updating in both visual and audio backbone networks in order to achieve best performance. 

More setting details are showcased in the supplement.

\begin{table}
	\tabcolsep=8.8pt
	\centering
	\caption{Quantitative Results of Sound Localization on VGG-SS}%
	\label{tab:compare_vggss}
	\begin{tabularx}{\linewidth}{@{}llcc} 
		\toprule
		Method    					        				&Pre-train       		&cIoU $\uparrow$       		&AUC $\uparrow$ \\
		\midrule\midrule
		\multicolumn{4}{l}{\emph{Training on VGG-Sound10k}}\\
		Attention \cite{Senocak18}$_{\text{CVPR}18}^{*}$    &Yes, InitParam+Signal  &0.160                 		&0.283 \\
		HardWay \cite{Chen21c}$_{\text{CVPR}21}^{*}$   	    &Yes, InitParam			&0.277  			   		&0.349 \\
		AVID \cite{Morgado21b}$_{\text{CVPR}21}^{*}$        &No	          			&0.276						&0.344 \\
		FNAC \cite{Sun23}$_{\text{CVPR}23}^{*}$				&Yes, InitParam+Signal  &\underline{0.317}	   	    &\underline{0.371} \\
		SSPL (w/o PCM)   									&Yes, InitParam			&0.253				   		&0.335 \\               
		SSPL (w/ PCM)    									&Yes, InitParam			&0.314						&0.369 \\
		SACL             					 				&Yes, InitParam+Signal  &\textbf{0.360}  			&\textbf{0.397} \\
		\midrule
		\multicolumn{4}{l}{\emph{Training on VGG-Sound144k}}\\
		Attention \cite{Senocak18}$_{\text{CVPR}18}^{*}$    &Yes, InitParam+Signal  &0.171                 		&0.287 \\
		AVObject \cite{Afouras20}$_{\text{ECCV}20}$  	    &No	          			&0.297 			       		&0.357 \\
		HardWay \cite{Chen21c}$_{\text{CVPR}21}^{*}$	    &Yes, InitParam			&0.319	   			   		&0.370 \\
		AVID \cite{Morgado21b}$_{\text{CVPR}21}^{*}$        &No	          			&0.315						&0.364 \\
		HardPos \cite{Senocak22}$_{\text{ICASSP22}}$		&Yes, InitParam+Signal  &0.346  			   		&0.380 \\
		TIE \cite{Liu22}$_{\text{ACM MM}22}$				&No	          			&\textbf{0.386}				&\underline{0.396} \\
		FNAC \cite{Sun23}$_{\text{CVPR}23}^{*}$				&Yes, InitParam+Signal  &0.357	   			        &0.388 \\
		SSPL (w/o PCM)   									&Yes, InitParam			&0.270	   			   		&0.348 \\
		SSPL (w/ PCM)    									&Yes, InitParam			&0.339	   	           		&0.380 \\
		SACL             					 				&Yes, InitParam+Signal  &\underline{0.380} 			&\textbf{0.400} \\
		\bottomrule
	\end{tabularx}%
\end{table}

\begin{figure}
	\centering
	\includegraphics[width=\linewidth]{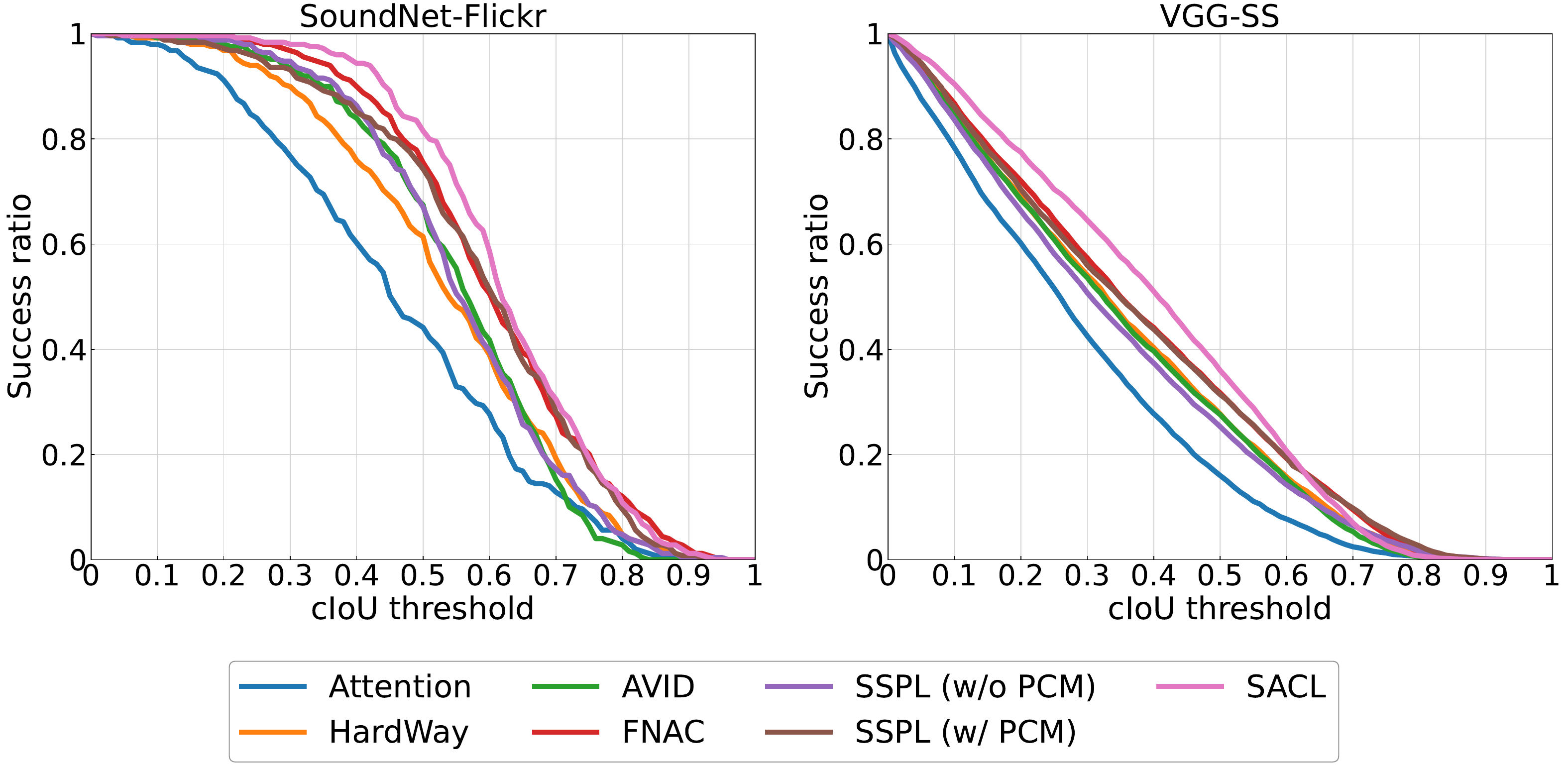}%
	\caption{Success ration with varying cIoU thresholds. Best viewed in color and by zooming in.}%
	\label{fig:success_rat_vs_ciou_thres}
\end{figure}

\begin{figure*}
	\centering
	\subfloat[Visualization on SoundNet-Flickr]{\includegraphics[width=0.5\linewidth]{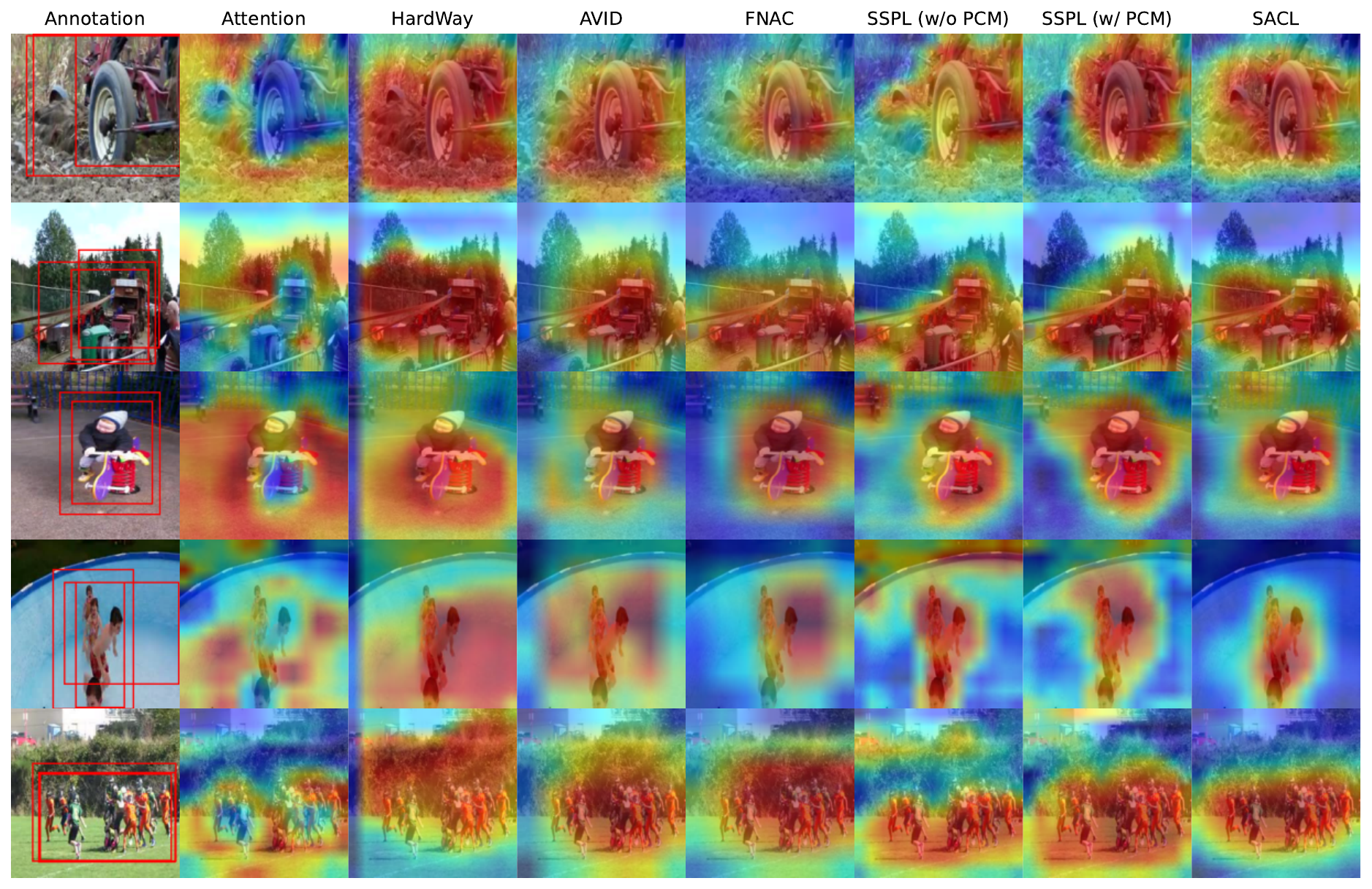}}%
	\subfloat[Visualization on VGG-SS]{\includegraphics[width=0.5\linewidth]{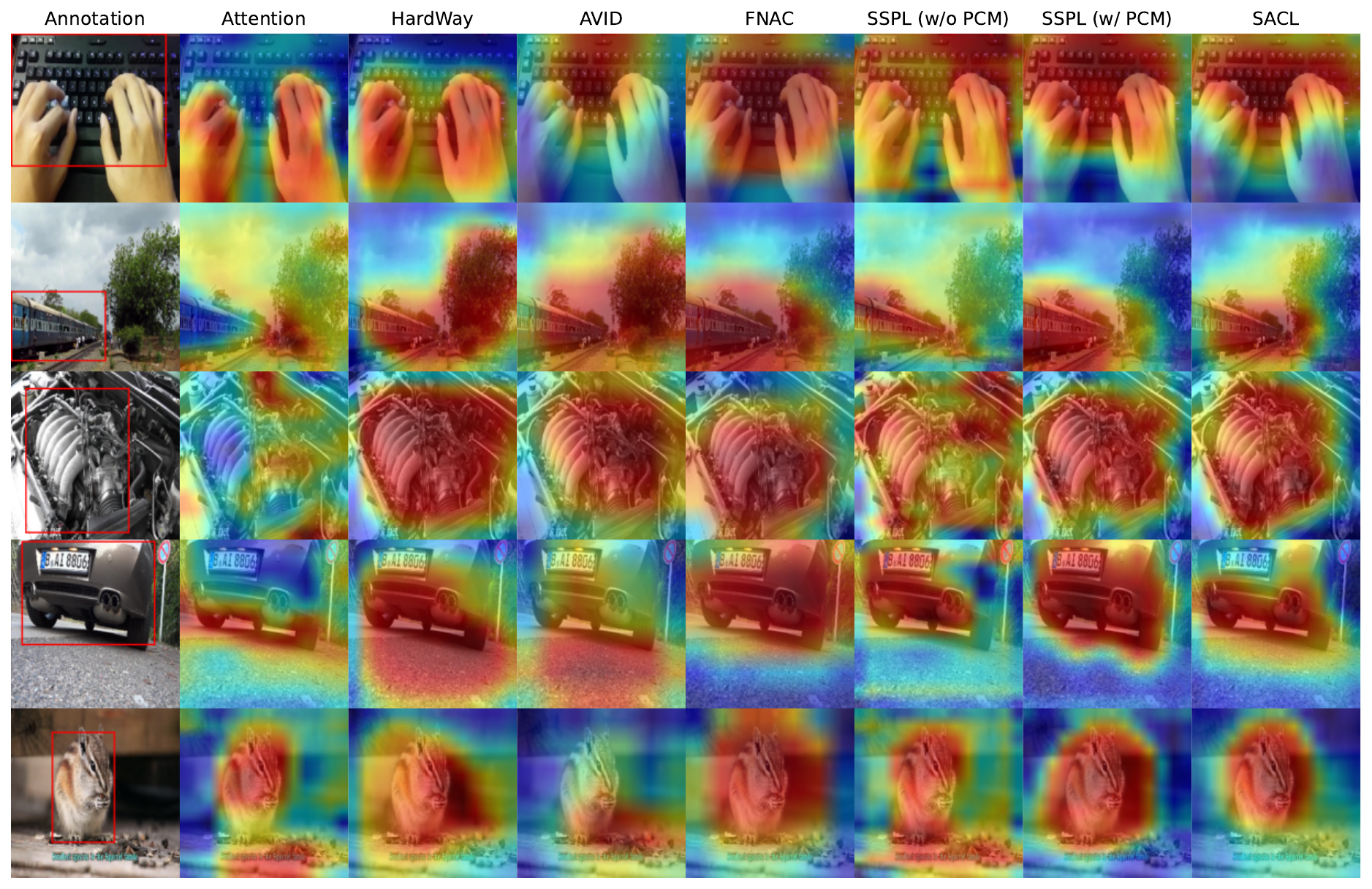}}%
	\caption{Qualitative comparisons. In each panel, the first column shows images accompanied with annotations, and remaining columns represent the predicted localization of sounding objects. Here the attention map or similarity map produced by different methods is visualized as the localization map. Note that for SoundNet-Flickr the bounding boxes are derived from multiple annotators.}%
	\label{fig:vis_att_map}
\end{figure*}

\subsection{Comparisons with State-of-the-art Methods}\label{sec:comp_sota}
\subsubsection{Results on SoundNet-Flickr}
We first compare SSPL and SACL with recent methods on the SoundNet-Flickr test set in Table \ref{tab:compare_flickr}. We find that (1) when trained by 10k Flickr samples, the vanilla SSPL (w/o PCM) performs favorably against HardWay, while showing inferior results compared with TIE and FNAC. However, the performance gap between SSPL and TIE is significantly reduced when learning with PCM (0.743 vs. 0.755 in cIoU and 0.587 vs. 0.588 in AUC). Similar phenomenon can be observed as training with 144k Flickr samples. These results verify the feasibility and effectiveness of our negative-free method for sound localization, and meanwhile suggest that PCM for feature alignment indeed facilitates the positive-only learning. (2) Our contrastive learning-based approach SACL outperforms all other competitors by a large margin. Particularly, in the Flickr144k training case, SACL increases performance by 4.4\% cIoU and 2.0\% AUC compared with TIE, establishing a new state-of-the-art on this benchmark. This illustrates the merit of audio-visual contrastive learning in terms of utilizing reliable anchor and negative features. (3) Following \cite{Chen21c}, we also train on VGG-Sound using respective 10k and 144k data pairs, which enables SSPL and SACL to achieve superior localization performance in both settings. As discussed in \cite{Chen21c}, the sounding objects are often visible in video clips from VGG-Sound, revealing that our method can benefit from the improved data quality. (4) Compared with SSPL (w/ PCM), SACL performs better when scaling up the size of training set (from Flickr10k to Flickr144k: 2.2\% improvement of cIoU for SSPL and 4.4\% for SACL; from VGG-Sound10k to VGG-Sound144k: 0.5\% improvement of cIoU for SSPL and 3.0\% for SACL). This is partly because large-scale dataset contains abundant semantics in both audio and visual modalities; such semantics could offer various negatives for SACL to regularize the representation space; however, SSPL cannot learn with this regularization due to its negative-free nature.

\subsubsection{Results on VGG-SS}
We further evaluate our approach on the newly released VGG-SS dataset and report results in Table \ref{tab:compare_vggss}. Because in this challenging benchmark the sounding object categories are more diverse and the number of test samples is greater than those of SoundNet-Flickr \cite{Chen21c}, the performances of all methods drop severely compared with those in Table \ref{tab:compare_flickr}. While SSPL (w/o PCM) still outperforms Attention by a large margin, it does not overtake HardWay. We attribute this to the limitation of vanilla SSPL on suppressing background noise (see Section \ref{sec:qual_analysis} for an empirical comparison). However, by combining with feature alignment module, SSPL (w/ PCM) yields performance better than HardWay (0.339 vs. 0.319 in cIoU, over 6\% gain) and on par with HardPos (0.339 vs. 0.346 in cIoU) in the 144k's scenario. This again demonstrates the advantage of the enhanced SSPL. Additionally, SACL appears to achieve performance superior or equivalent to the concurrent best art TIE (the difference between cIoUs is 0.006, and 0.004 for AUCs). However, our method proceeds without using any audio augmentations, which prove to be crucial for TIE \cite{Liu22}.

\subsubsection{Generalization with Various cIoU Thresholds}
To address diverse demands for sound localization fineness, we compute cIoU scores with different thresholds as shown in Fig. \ref{fig:success_rat_vs_ciou_thres}. The proposed method, SSPL (w/ PCM), again consistently surpasses Attention and HardWay under all thresholds. Besides, SACL shows better localization performance than SSPL (w/ PCM) across all thresholds on SoundNet-Flickr and across thresholds less than 0.63 on VGG-SS, illustrating that SACL can capture meaningful regions of sounding objects in most cases.

\subsection{Qualitative Analysis}\label{sec:qual_analysis}
\subsubsection{Visualization of Localization Map}
We provide visualized localization results in Fig. \ref{fig:vis_att_map}. We observe that Attention is prone to overlook target objects (e.g., the first and second rows in Fig. \ref{fig:vis_att_map}a) and cover unrelated background details (e.g., ground and sky). Since localization map also visualizes similarities between audio and visual features, the inaccurate localization indicates that Attention has the undesired potential to misalign features. Although HardWay presents more centralized attention via hard negative mining, it easily underestimates (e.g., the first row in Fig. \ref{fig:vis_att_map}b) or overestimates (e.g., the second row in Fig. \ref{fig:vis_att_map}b) extents of sounding objects. This is probably because positive and negative regions in different images cannot be simply distinguished by the same thresholding parameters \cite{Chen21c}. The localization maps of AVID exhibit the same problems like HardWay. By contrast, FNAC and our SSPL can almost cover the entire region of interest, and the use of PCM further helps SSPL reduce the influence of background noise, leading to more accurate localization. As for SACL, its attention is largely concentrated on sounding objects and is less emphasized on irrelevant distractions (e.g., the penultimate row in Fig. \ref{fig:vis_att_map}a and the last row in Fig. \ref{fig:vis_att_map}b), resulting in the most compact localization map against all other competitors. We have also shown the localization results on videos, which can be found in the supplement.

\subsubsection{Learned Audio-Visual Correspondence}
\begin{figure}
	\centering
	\includegraphics[width=0.95\linewidth]{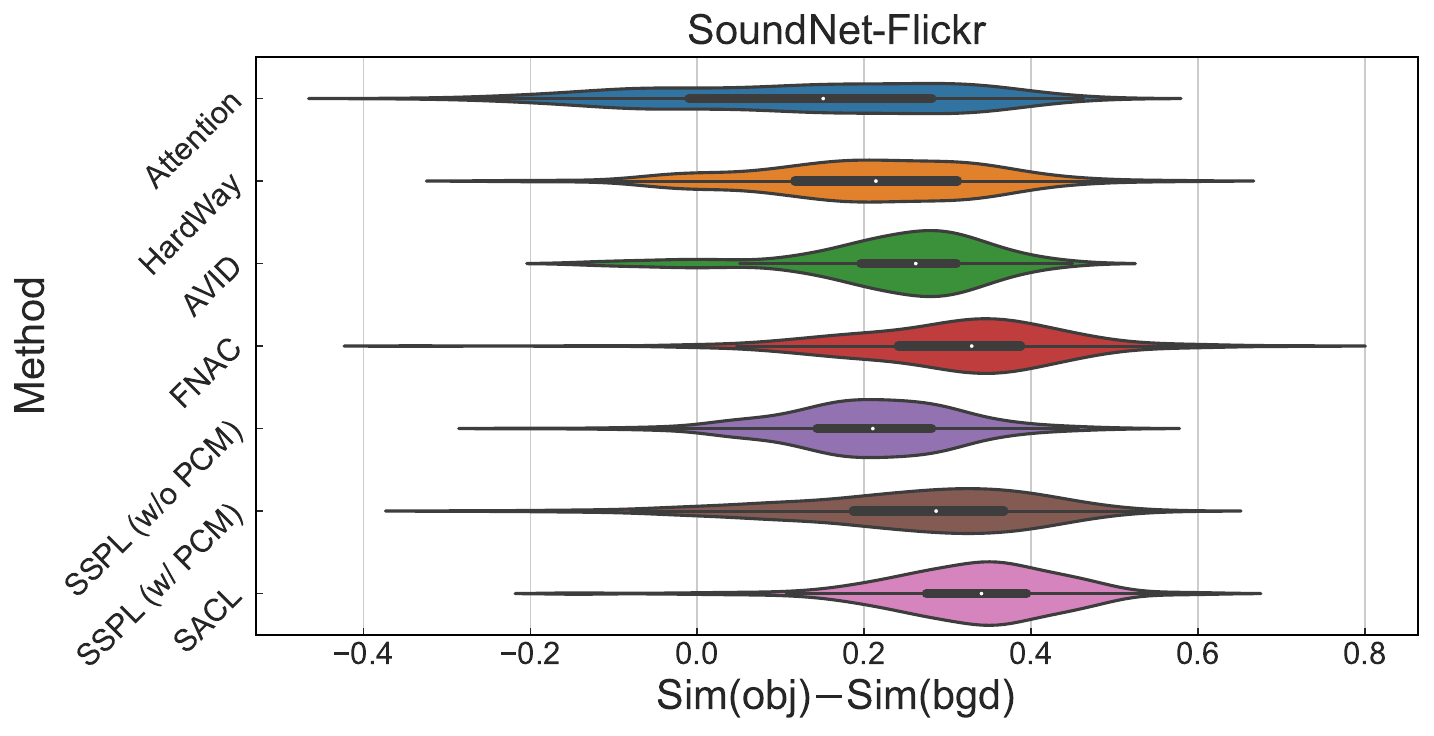}\\
	\vspace{2mm}
	\includegraphics[width=0.95\linewidth]{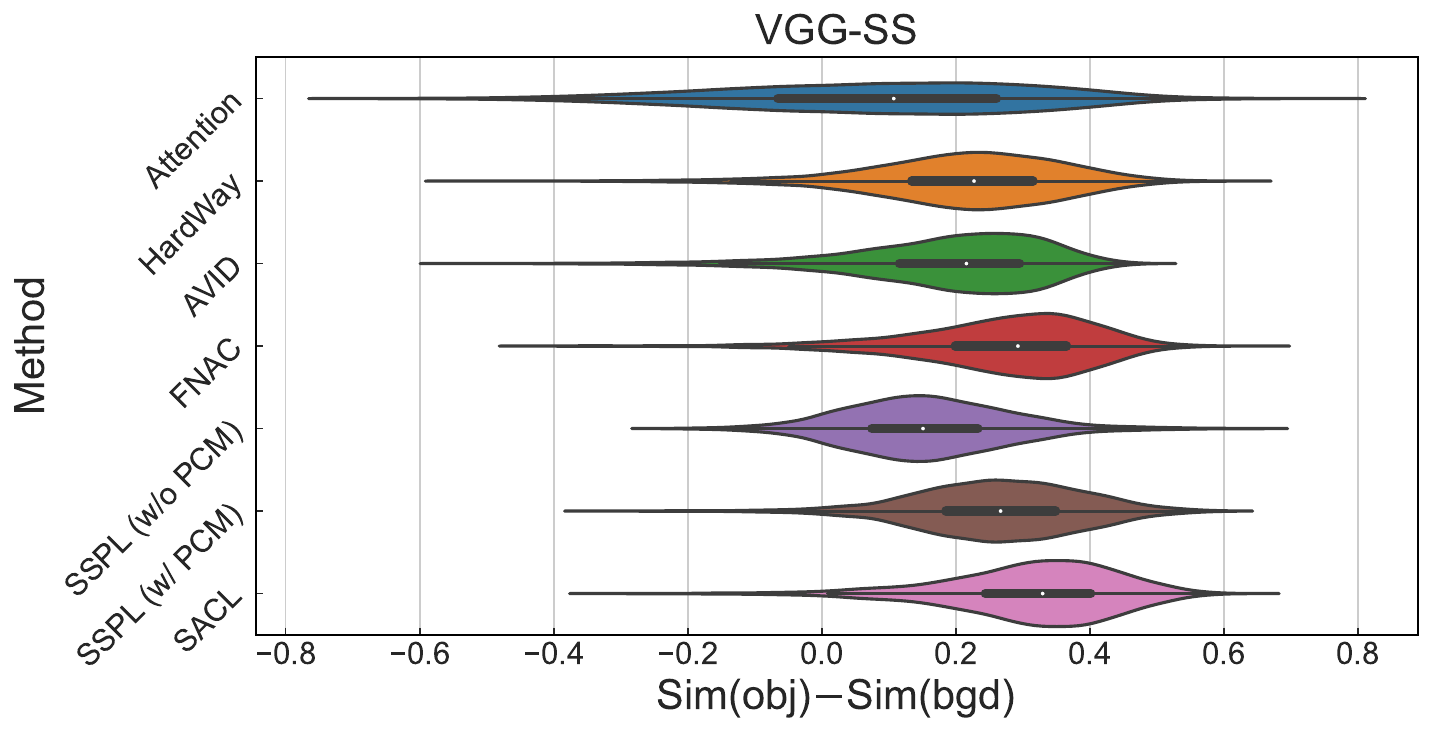}%
	\caption{Distribution of the learned audio-visual similarity. The label on horizontal axis, ``Sim(obj)$-$Sim(bgd)'', denotes the difference between audio-visual similarities computed in object region and background region (here, we use the normalized similarity in Eq. \eqref{eq:norm_sim_map} for fair comparisons). The greater the similarity difference, the more accurate the learned correspondence.}
	\label{fig:dist_sim_diff}
\end{figure}

Learning precise audio-visual correspondence is the core objective for sound localization. This requires audio and visual features of the same sounding object to be as similar as possible, while as dissimilar as possible if they are from different objects. To analyze this property, we compute the difference of the averaged audio-visual similarity between sounding object region and background region, and further visualize the distribution of such similarity difference in Fig. \ref{fig:dist_sim_diff}. It can be observed that Attention outputs similarities that are not well distinguishable in object and background regions, indicating ambiguous correspondence. HardWay and AVID capture more accurate audio-visual correlation than SSPL (w/o PCM) on VGG-SS, but have no advantage vs. SSPL (w/ PCM) on both datasets. Note that as two types of false negative-aware sound localization methods, FNAC and SACL achieve most distinguishable similarities in different regions, which is also manifested by the mean of the similarity difference (i.e., 0.312 for FNAC vs. 0.330 for SACL on SoundNet-Flickr; 0.271 for FNAC vs. 0.313 for SACL on VGG-SS). These results validate the effectiveness of the negative-calibrated contrastive learning in aligning semantic audio and visual features.

\subsection{Ablation Study on SSPL}\label{sec:ablation_sspl}
In this section, we delve deeper into SSPL by conducting extensive ablation studies, unfolding what component matters in our negative-free learning scheme. Unless otherwise specified, all experiments are performed on SoundNet-Flickr.

\subsubsection{Training Strategy}\label{sec:ablation_sspl_train}
As discussed in prior art \cite{Chen21a}, a simple Siamese network without using negative samples can easily suffer from the problem of representation collapse. In this regard we evaluate key factors of SSPL that facilitate audio-visual learning. In Table \ref{tab:ablation_sspl_train}a we train the model from scratch while removing the stop-gradient operation, which indeed causes collapse in our practice. The variant with only pre-training strategy (Table \ref{tab:ablation_sspl_train}b) improves performance because of the better parameter initialization, but it dose not avoid collapsed solution yet. Adding stop-gradient alone during training (Table \ref{tab:ablation_sspl_train}c) can obtain obvious gains, and the combination with pre-training (Table \ref{tab:ablation_sspl_train}d) further boosts cIoU to $0.671$, which is the default configuration of vanilla SSPL. 

Based on above configuration, we perform additional ablation on the recursive cycles for representation updates in PCM. The performance sightly drops by 2\% when conducting feedback and feedforward representation updates (Eqs. \eqref{eq:feedback_pred}-\eqref{eq:feedforward_repres}) only once (Table \ref{tab:ablation_sspl_train}e). This is because one computing step is not enough for PCM to reduce prediction errors between audio and visual features, and such non-negligible errors could degrade the subsequent learning. However, by increasing recursive cycles, SSPL can harvest significant performance improvements (nearly 10\% in Table \ref{tab:ablation_sspl_train}f and over 13\% in Table \ref{tab:ablation_sspl_train}g, respectively).

In summary, the results demonstrate that stop-gradient also works in our audio-visual setting to prevent collapse; and that both pre-training and PCM induce the model to learn effectively so as to promote localization accuracy.

\begin{table}
	\tabcolsep=10.9pt
	\centering
	\caption{Ablation on Training Strategies of SSPL}%
	\label{tab:ablation_sspl_train}
	\begin{threeparttable}
		\begin{tabularx}{\linewidth}{@{}*{6}{c}}
			\toprule
			{}            &IN-Pre-train        &Stop-grad       &$T$         &cIoU $\uparrow$   		&AUC $\uparrow$ \\
			\midrule\midrule
			(a) 	      &{}				   &{}			    &{}	 	     &0.141            			&0.147 \\
			(b) 	      &$\checkmark$		   &{}			    &{}	 	     &0.382            			&0.432 \\
			(c)  	      &{} 	               &$\checkmark$	&{}	 	     &0.570            			&0.511 \\
			(d)   	   	  &$\checkmark$ 	   &$\checkmark$	&{}	 	     &\textbf{0.671} 			&\textbf{0.556} \\
			\midrule
			(e)   	   	  &$\checkmark$ 	   &$\checkmark$ 	&1           &0.655            			&0.562 \\
			(f)   	   	  &$\checkmark$ 	   &$\checkmark$ 	&3 		     &0.719            			&0.584 \\
			(g)   	   	  &$\checkmark$ 	   &$\checkmark$ 	&5 		     &\textbf{0.743} 			&\textbf{0.587} \\
			\bottomrule
		\end{tabularx}%
		\begin{tablenotes}[para,flushleft]
			\emph{``IN-Pre-train'' represents using the ImageNet-pre-trained backbone to extract visual features, and $T$ denotes the recursive cycles for iterative computing in PCM during training.}
		\end{tablenotes}%
	\end{threeparttable}
\end{table}

\subsubsection{Scaling Method in Attention Module (AM)}\label{sec:ablation_sspl_scaling}
\begin{table}
	\tabcolsep=33pt
	\centering
	\caption{Ablation on Scaling Methods in Attention Module (AM)}%
	\label{tab:ablation_sspl_scaling}
	\begin{tabularx}{\linewidth}{@{}l*{2}{c}}
		\toprule
		Scaling method          		 &cIoU $\uparrow$      		&AUC $\uparrow$ \\
		\midrule\midrule
		ReLU \cite{Qian20}      		 &0.353				   		&0.424 \\
		Sigmoid				    		 &0.647				   		&0.547 \\
		Softmax \cite{Senocak18}		 &0.667				   		&0.554 \\
		ReLU + Softmax \cite{Senocak18}	 &0.574				   		&0.531 \\
		Min-Max Norm.	        		 &\textbf{0.671}			&\textbf{0.556} \\
		\bottomrule
	\end{tabularx}%
\end{table}

\begin{table*}
	\tabcolsep=14.8pt
	\centering
	\caption{Ablation on Training Strategies of SACL}%
	\label{tab:ablation_sacl_train}
	\begin{threeparttable}
		\begin{tabularx}{\linewidth}{@{}*{9}{c}}
			\toprule
			\multirow{2}{*}{}  &\multirow{2}{*}{Pre-train} &\multirow{2}{*}{Fine-tune} &\multirow{2}{*}{Pseudo mask}  &\multirow{2}{*}{\thead{False negative \\ detection}} &\multicolumn{2}{c}{SoundNet-Flickr} 		&\multicolumn{2}{c@{}}{VGG-SS} \\
			\cmidrule(lr){6-7}\cmidrule(lr){8-9}
							   &{}          			   &{}					       &{}				   			  &{}				     								&cIoU $\uparrow$ 	&AUC $\uparrow$     	&cIoU $\uparrow$    &AUC $\uparrow$ \\
			\midrule\midrule
			(a) 	           &$\checkmark$	    	   &{}	                       &{}			       			  &{}	 	        	 								&0.619           	&0.525 		     		&0.265           	&0.351 \\
			(b) 	           &$\checkmark$	    	   &$\checkmark$               &{}			       			  &{}	 	        	 								&0.707           	&0.566 		     		&0.341           	&0.386 \\
			(c)   	   	       &$\checkmark$        	   &$\checkmark$               &$\checkmark$		   	      &{}   		         								&0.755   		 	&0.593 		     		&0.349           	&0.388 \\
			(d)  	           &$\checkmark$        	   &$\checkmark$	           &{}	      		   			  &$\checkmark$	 	 									&0.763           	&0.585 		     		&0.355           	&0.395 \\
			(e)   	   	       &$\checkmark$        	   &$\checkmark$               &$\checkmark$         		  &$\checkmark$   		 								&\textbf{0.815} 	&\textbf{0.623} 	 	&\textbf{0.360}  	&\textbf{0.397} \\
			\bottomrule
		\end{tabularx}%
	\end{threeparttable}
\end{table*}

The similarity map takes values in $[-1,1]$ and is adapted to weigh visual features in AM. Here we study different methods that can scale similarity range into $[0,1]$. ReLU is used in \cite{Qian20} to compact the similarities less than 0, but in our model enforcing those negative values to be equal produces worst results, as shown in Table \ref{tab:ablation_sspl_scaling}. While sigmoid and softmax \cite{Senocak18} boost performance by taking all different similarities into account, they shrink values into a proper subset of $[0,1]$. The min-max normalization (Eq. \eqref{eq:norm_sim_map}), by contrast, takes a step forward and separates minima and maxima to the largest extent, yielding best results among others. This reveals that the relative importance between spatial-wise visual features is more crucial than the feature value per se for sound localization task.

\begin{figure}
	\centering
	\includegraphics[width=0.48\linewidth]{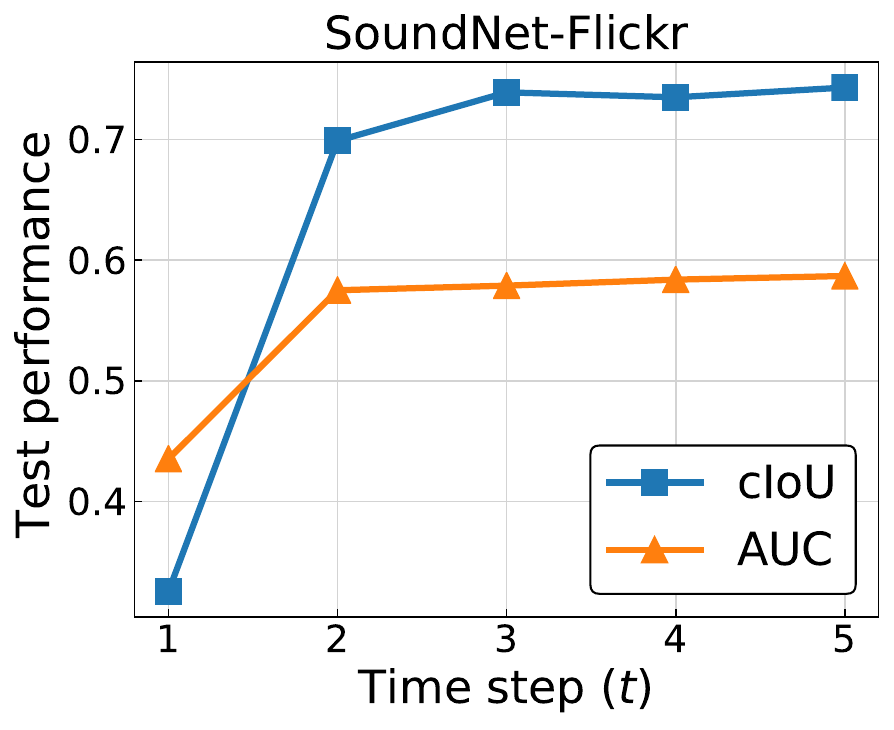}\hspace{2.7mm}
	\includegraphics[width=0.48\linewidth]{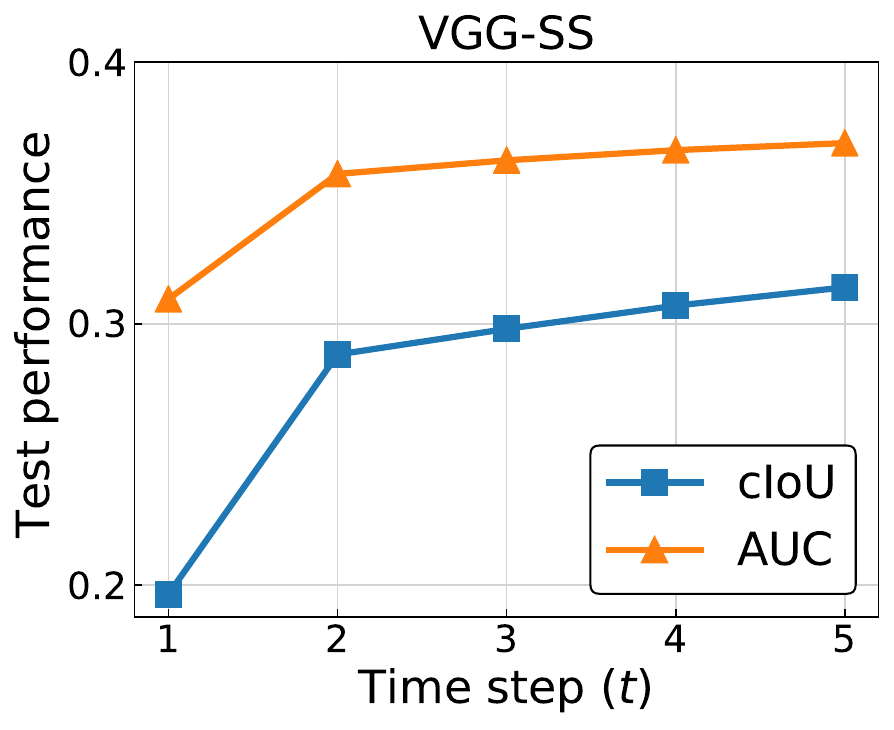}%
	\caption{Performance with PCM's iterations during test.}%
	\label{fig:perform_vs_cycles}
\end{figure}

\begin{figure}
	\centering
	\includegraphics[width=\linewidth]{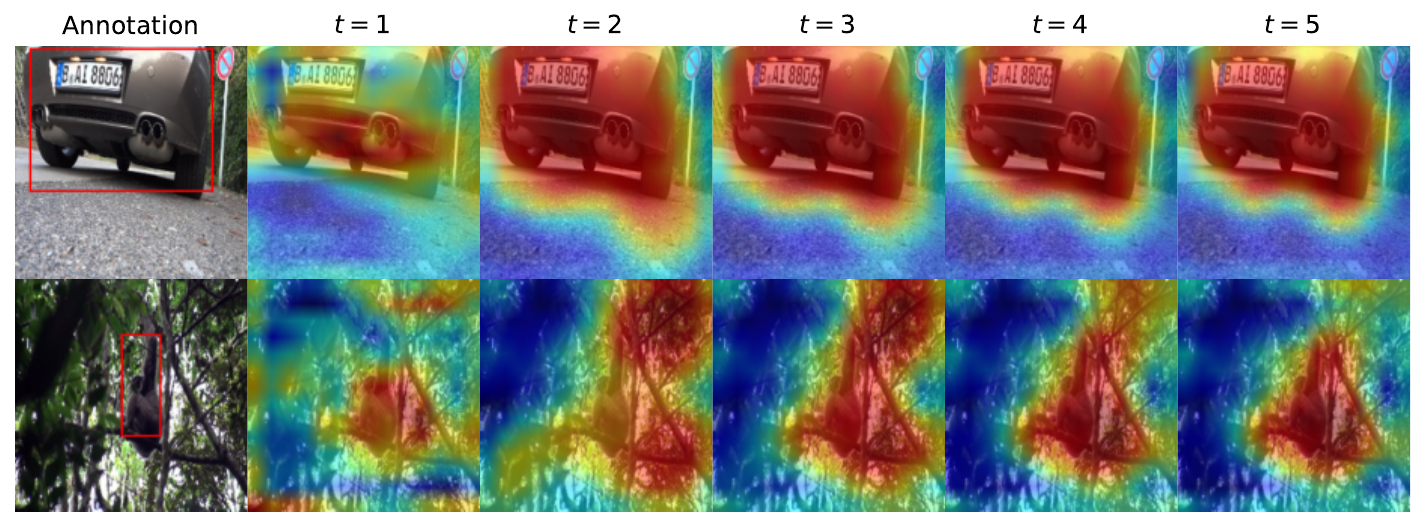}%
	\caption{Attention map with PCM's iterations during test. Illustrations are from VGG-SS.}%
	\label{fig:vis_att_map_vs_iters}
\end{figure}
\subsubsection{Further Analysis of Predictive Coding Module (PCM)}\label{sec:ablation_sspl_pcm}
We empirically clarify the remarkable ability of PCM to boost sound localization. Since PCM features an iterative computing procedure, we inspect the performance of SSPL (w/ PCM) with different iterations. From Fig. \ref{fig:perform_vs_cycles}, we see that the localization accuracy tends to increase given more iterative computations, especially at the initial three time steps. To understand why this is the case, we look into attention maps from some test samples, as illustrated in Fig. \ref{fig:vis_att_map_vs_iters}. PCM infers different visual representations with varying time steps (1 through 5), which are further used by AM to produce different attention maps. Attention is less definitive (light red on sounding objects) and/or inaccurate (crimson on background) at early time steps. At later time steps, however, the model corrects itself to pay more definitive and accurate attention to sounding objects. Adjusting attention in such a coarse-to-fine way is particularly helpful to address ambiguous cases, where the object's appearance may be similar to background (e.g., the second row in Fig. \ref{fig:vis_att_map_vs_iters}).

\subsection{Ablation Study on SACL}\label{sec:ablation_sacl}
We provide a detailed ablation study to demonstrate how each component in SACL contributes to the performance. We first analyze the effect of different training strategies on localization. Then, we quantitatively compare the spatial heuristic and image-computable masks used to compact visual features. At last, we illustrate the critical influence of false negatives.

\subsubsection{Training Strategy}\label{sec:ablation_sacl_train}
Table \ref{tab:ablation_sacl_train} summarizes the ablation results about different training settings in SACL. For simplicity, all models considered here are trained with 10k image-audio pairs. We regard the plain model initialized with pre-trained feature extractors as our baseline (Table \ref{tab:ablation_sacl_train}a). Compared with the plain model, fine-tuning both visual and audio backbone networks (Table \ref{tab:ablation_sacl_train}b) yields noteworthy performance improvement (0.707 vs. 0.619 in cIoU on SoundNet-Flickr). The result indicates the significant difference between classification task and sound localization task. SACL in Table \ref{tab:ablation_sacl_train}c further benefits from the pseudo mask used for compacting visual features (0.755 vs. 0.707 in cIoU on SoundNet-Flickr, nearly $7\%$ improvements), confirming the efficacy of reducing ambiguity in visual modality. In addition, SACL can also get $8\%$ gains with false negative detection (i.e., selectively retaining $k=N\times75\%$ negatives in a batch, Table \ref{tab:ablation_sacl_train}d) compared with the variant in Table \ref{tab:ablation_sacl_train}b. This is because removing potential false negatives enables contrastive learning to capture unambiguous cross-modal relation in semantic feature space. Finally, the combination of two individual components, pseudo mask and false negative detection, results in the best localization performance over all variants (Table \ref{tab:ablation_sacl_train}e). We can draw the similar conclusion from the results on VGG-SS.

\subsubsection{Unsupervised Mask Generation}\label{sec:ablation_sacl_mask}
\begin{figure}
	\centering
	\includegraphics[width=\linewidth]{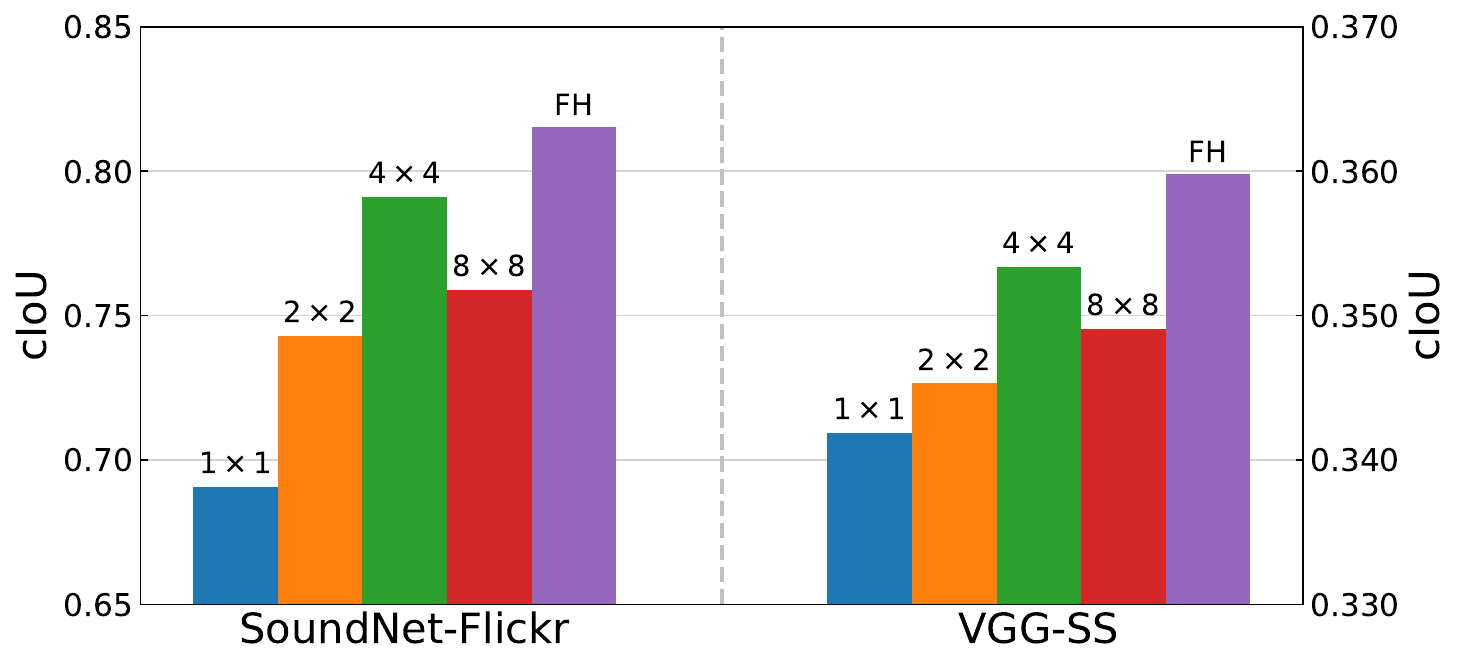}\\
	\caption{Effect of the type of masks used in SACL. Compared with spatial grid masks ($d\times d$), the image-computable FH mask enables SACL to produce more accurate localization.}
	\label{fig:effect_mask_type}
\end{figure}

The SACL can be instantiated with various image segmentation strategies. Here we compare two types of them, i.e., spatial heuristics and image-computable FH algorithm, by inspecting the resultant performance as adopting different pseudo masks (example masks are illustrated in Fig. \ref{fig:pseudo_mask_eg}). As shown in Fig. \ref{fig:effect_mask_type}, the version of using $1\times1$ mask is equivalent to compacting visual features with only similarity map, and is viewed as a baseline solution. SACL equipped with $4\times 4$ grid mask contrasts audio features and visual features inside the one-sixteenth region of the visual feature map. Although this spatial decomposition is simple, it outperforms the baseline by a large margin, showing promise of audio-visual contrastive learning at fine-grained level. However, finer grids cannot improve performance, since smaller regions may contain too fewer visual features to be adequate for learning. By contrast, the FH algorithm assists SACL to achieve the best localization. We attribute such superiority to FH's flexible way to compute object-based masks, which in fact provide semantic-aware candidate regions to filter visual features of interest.

\subsubsection{Effect of False Negatives}\label{sec:ablation_sacl_false_neg}
\begin{figure}
	\centering
	\includegraphics[width=\linewidth]{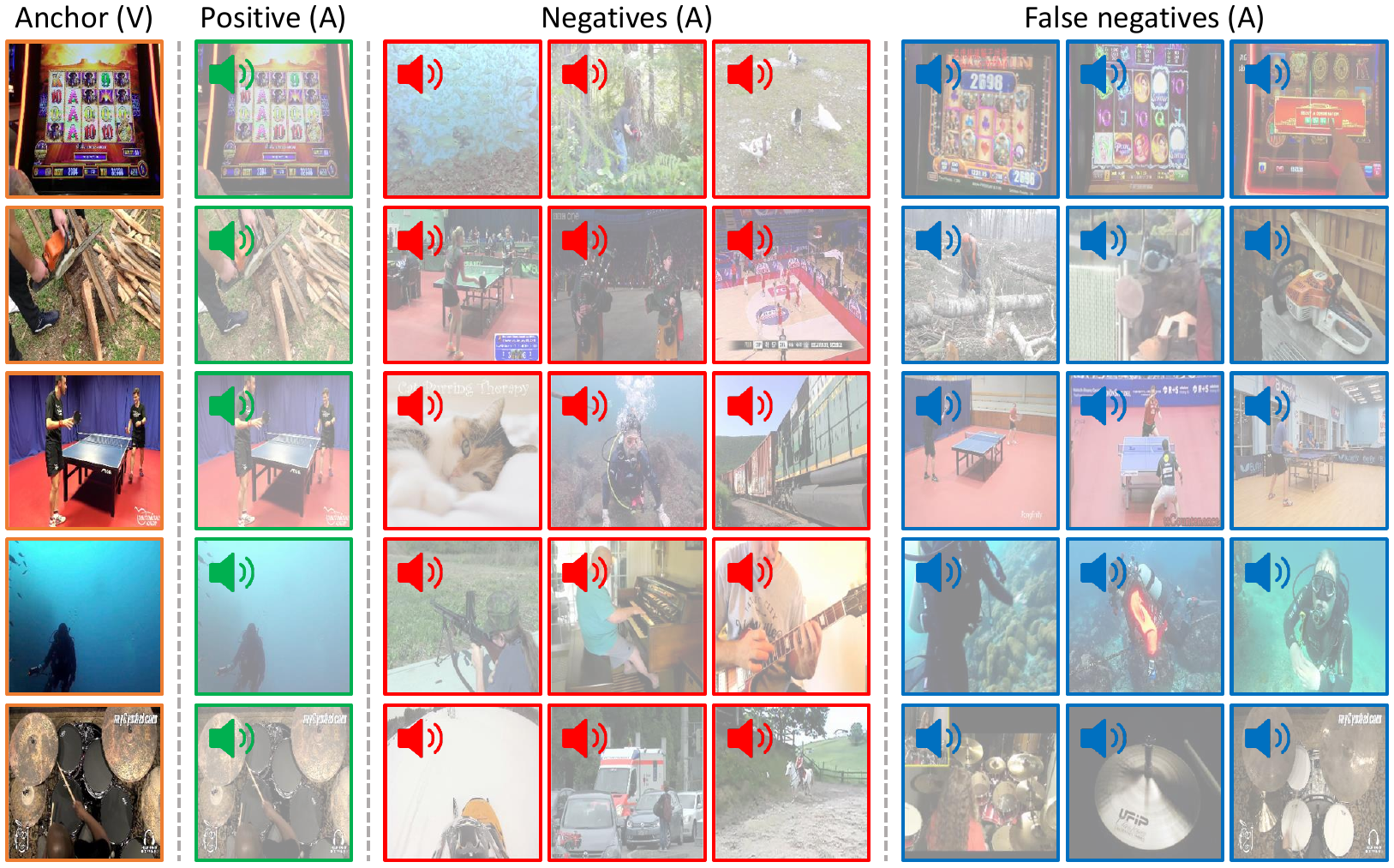}%
	\caption{Illustration of the VGG-Sound144k training samples selected by SACL. ``V'' and ``A'' denote visual and audio modalities, respectively. The whitened image with sound icon indicates the semantic content of the audio. For each visual anchor, we respectively show its positive sample, three negatives in the constructed negative set, and three false negatives eliminated due to high audio feature similarity to the positive.}%
	\label{fig:vis_anchor_pos_neg_false_neg}
\end{figure}

We first investigate the performance of SACL in false negative detection. For all compared methods, we calculate the intra-modality feature similarity between a batch of training data. For each anchor sample, a quarter of batch data with highest similarities are viewed as false negative candidates. As exhibited in Table \ref{tab:detect_false_neg}, FNAC consistently detects more false negatives than the audio classification method VGGish ($78\%$ vs. $44\%$), showing the superiority of multimodal learning in recognizing semantic concepts of different classes. By learning audio-visual representations based on the pre-trained VGGish, SACL endows extracted audio features with semantic discriminability. Accordingly, SACL identifies the most false negatives in all settings ($88\%$). We further visualize some training data sampled by SACL in Fig. \ref{fig:vis_anchor_pos_neg_false_neg}, which indicates that SACL can distinguish reliable negatives from false negatives based on the audio feature similarity.

\begin{figure*}
	\centering
	\subfloat[Sampling negatives randomly or selectively]{
		\includegraphics[width=0.48\linewidth]{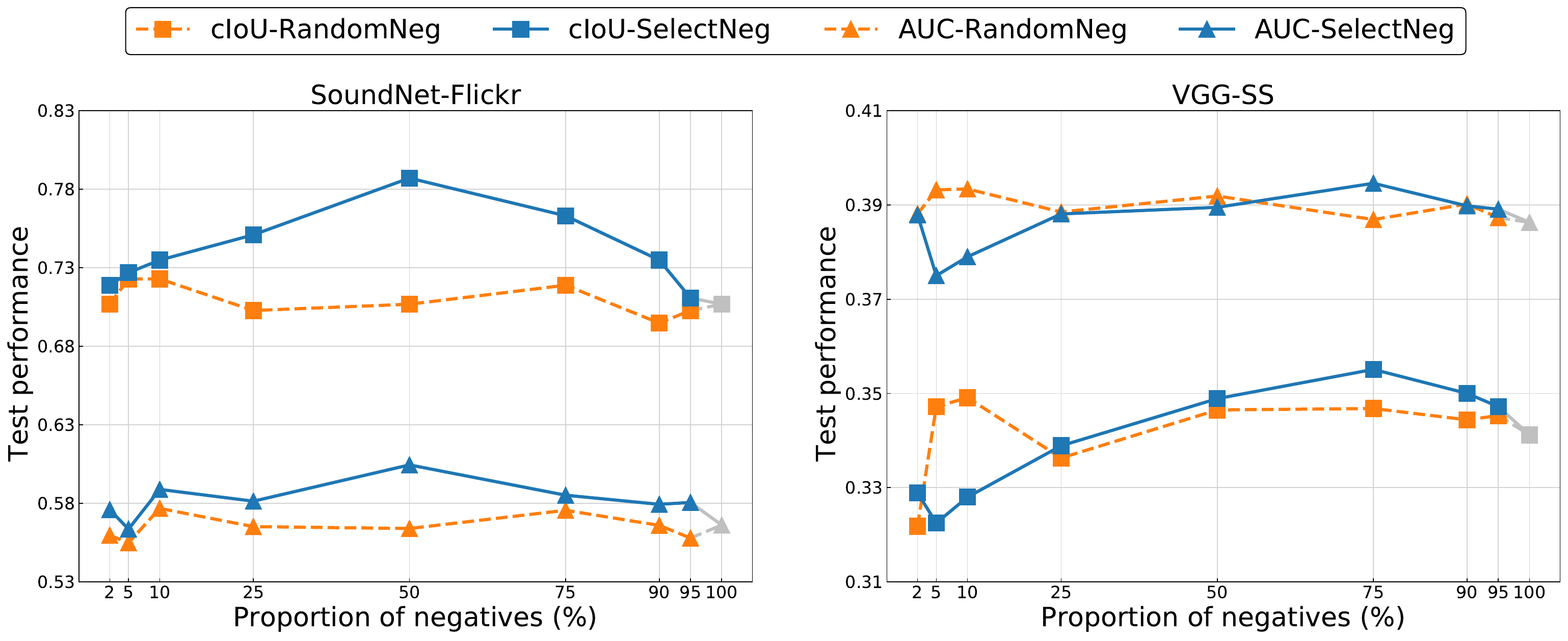}
	}\hspace{2.5mm}
	\subfloat[Performance with varying proportions of negatives]{
		\includegraphics[width=0.48\linewidth]{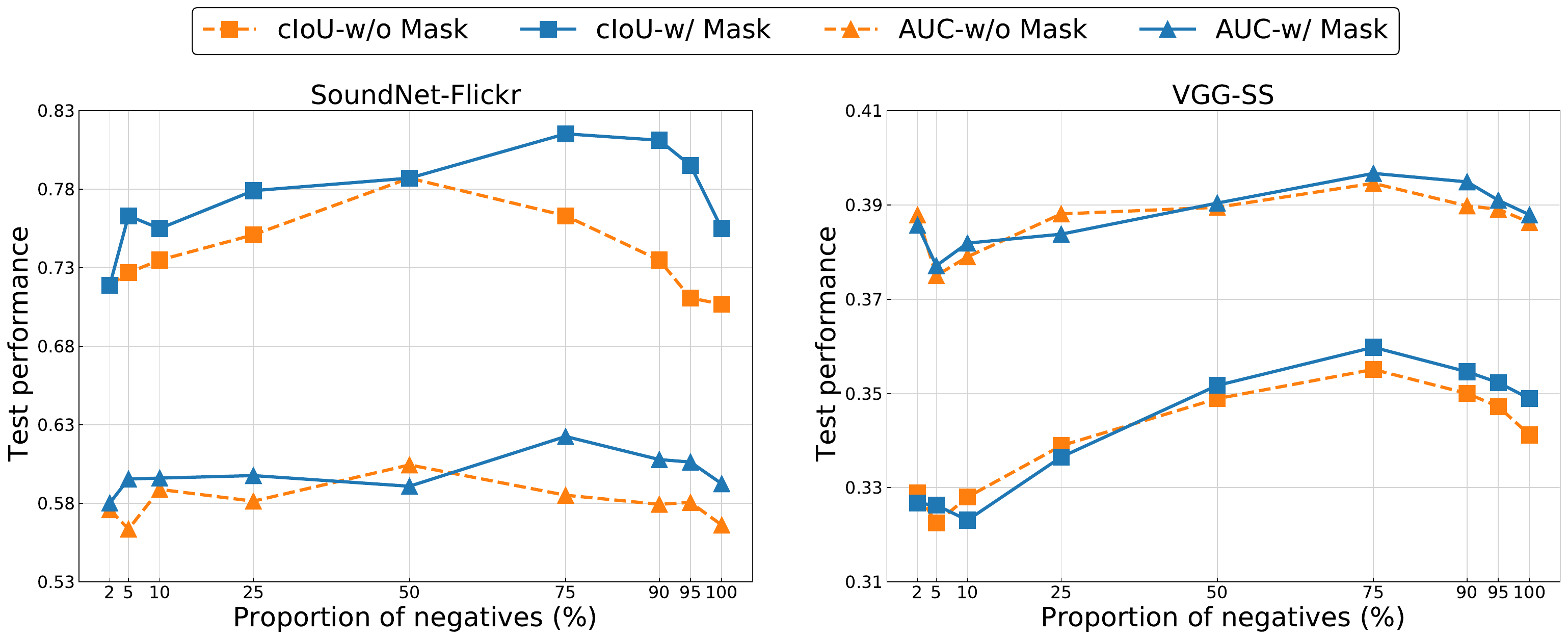}
	}
	\caption{Effect of false negatives in terms of sampling strategy and sampling proportion. In both (a) and (b), we sample different proportions of audio negatives within a mini-batch for contrastive learning. The proportion $2\%$ refers to only sampling one negative (when batch size is 64 on each GPU), and $100\%$ (marked in {\color{gray}\textbf{gray}} in (a)) means viewing all audio samples extracted from different videos as negatives. Note that the basic SACL used in (a) excludes the visual feature compaction for simplicity.}%
	\label{fig:effect_false_neg}
\end{figure*}

We then quantitatively illustrate the effect of false negatives on sound localization regarding sampling strategy. Fig. \ref{fig:effect_false_neg}a compares two sampling strategies for obtaining negatives: one is \emph{random sampling} (as used in most existing works), which considers all sounds of videos different from the given one as negative samples, and randomly extracts one or more sounds from them for contrastive learning; the other is the proposed \emph{selective sampling}, which treats sounds with low audio feature similarity to the given sound source as negatives. We find that (1) the random strategy achieves competitive performance when drawing only a few negatives ($\leqslant10\%$ batch size), but additional improvements cannot be harvested as increasing the proportion. This is because in the random sampling scenario, larger proportion of negatives are more likely to contain false negatives, which would prevent contrastive learning from building cross-modal correspondence. (2) Our practice avoids this issue via explicitly detecting and removing false negatives (i.e., using FND), hence more reliable negatives are incorporated for contrast with the increase in proportion. As a result, the selective sampling performs favorably against the random counterpart, especially when the number of sampled negatives is more than half of the batch size (i.e., proportion $\geqslant50\%$). (3) It is worth mentioning that taking full advantage of all potential negatives (i.e., proportion $=100\%$) yields inferior results compared with our selective strategy. This also highlights the necessity to utilize semantically dissimilar negatives for audio-visual contrastive learning.

\begin{table}
	\tabcolsep=11pt
	\centering
	\caption{Performance Evaluation of False Negative (FN) Detection on VGG-Sound144k Training Set (for A Batch of Data on Average)}%
	\label{tab:detect_false_neg}
	\begin{threeparttable}
		\begin{tabularx}{\linewidth}{@{}l*{5}{c}}
			\toprule
			\multirow{2}{*}{Method}       				&\multicolumn{5}{c}{Batch size}  \\
			\cmidrule(lr){2-6}
			&64     		&128        	&256        	&512        	&1024     	   \\
			\midrule\midrule
			No. of FN         							&17				&68				&271			&1082			&4319	  	   \\
			\midrule
			VGGish \cite{Hershey17}$_{\text{ICASSP}17}$ &7	    		&30				&123			&496			&1979	  	   \\
			FNAC \cite{Sun23}$_{\text{CVPR}23}$			&13				&53				&213			&855			&3423	  	   \\
			SACL 										&\textbf{15}	&\textbf{60}	&\textbf{240}	&\textbf{958}	&\textbf{3827} \\
			\bottomrule
		\end{tabularx}%
	\end{threeparttable}
\end{table}

We finally carry out experiments to determine the optimal proportion to sample negatives. Fig. \ref{fig:effect_false_neg}b shows the results of two variants of SACL: embracing or abandoning pseudo masks. We can see that the performance of both variants is gradually boosted by increasing the proportion, and best results are reached at some specific proportion, i.e., $50\%$ for SACL without using mask on SoundNet-Flickr and $75\%$ for all other cases. This performance trend from rise to decline is expected, because more negatives selected effectively are beneficial for contrastive learning; and meanwhile, false negatives will naturally emerge when the sampling proportion gets closer to $100\%$. We fix the proportion to $75\%$ across all other experiments in this paper. One can also notice that the performance gap between the two SACLs becomes large as more than half of the potential negatives are selected. This is mainly because when learning without pseudo masks, audio negatives might be contrasted with the visual feature from background distractions. Duo to the incorrect visual anchor features, including more negatives would, on the contrary, improve the probability of feature misalignment. By contrast, compacting visual features by pseudo mask enables SACL to focus on meaningful visual anchors, and thus learning could benefit from abundant negatives.

\section{Generalizability to other Tasks}
\subsection{Audio-Visual Event Classification}
In order to verify the generalizability of the proposed scheme, we first evaluate the representation quality on the audio-visual event classification task. Specifically, we keep the encoders of the pre-trained audio-visual models and connect them to a three-layer MLP head. We use a smaller learning rate $10^{-5}$ to fine-tune the pre-trained weights while a $10\times$ larger learning rate for the MLP head. All methods are compared by fine-tuning the corresponding model with audio-only data, visual-only data, and audio-visual data from the VGG-Sound144k dataset, respectively.

We show classification results on VGG-SS in Table \ref{tab:av_classification}. For all methods, the classification accuracy derived from the audio-visual representation is considerably higher than those from unimodal representations, demonstrating the advantage of combining unimodal information to improve semantic discriminability. Besides, compared with the methods using random sampling strategy during audio-visual pre-training (i.e., Attention and HardWay), the rest that are able to alleviate the false negative problem show consistent superiority when classifying based on audio-visual representation. This phenomenon indicates that learning multimodal representations with semantically unambiguous positives and negatives can prompt the downstream classification task. While SACL performs on par with SSPL's in the unimodal setting, it achieves the highest accuracy for the multimodal case. This reveals the benefit of explicitly eliminating false negatives in audio-visual correspondence learning.
\begin{table}
	\tabcolsep=14.5pt
	\centering
	\caption{Audio-Visual Event Classification Results (Accuracy in $\%$) on VGG-SS}%
	\label{tab:av_classification}
	\begin{threeparttable}
		\begin{tabularx}{\linewidth}{@{}l*{3}{c}}
			\toprule
			Method 			                            		&Audio       		    	&Visual     			    &Audio-Visual \\
			\midrule\midrule
			Attention \cite{Senocak18}$_{\text{CVPR}18}$        &38.36						&41.07						&41.88 \\
			HardWay \cite{Chen21c}$_{\text{CVPR}21}$   	        &36.96						&31.48						&43.50 \\
			AVID \cite{Morgado21b}$_{\text{CVPR}21}$            &31.42						&33.42						&43.93 \\
			FNAC \cite{Sun23}$_{\text{CVPR}23}$                 &33.53						&43.90						&\underline{54.41} \\
			SSPL (w/o PCM)   									&38.15						&\underline{45.42}			&49.83 \\               
			SSPL (w/ PCM)										&\underline{39.34}			&\textbf{45.61}				&52.58 \\
			SACL       											&\textbf{39.45}				&45.27						&\textbf{56.39} \\
			\bottomrule
		\end{tabularx}%
	\end{threeparttable}
\end{table}

\begin{table*}
	\tabcolsep=8.1pt
	\centering
	\caption{Object Detection Performance ($\text{AP}_{50}$ in $\%$ for Randomly Selected 10 Classes and $\text{mAP}_{50}$ in $\%$ over All 215 Classes) on the Subset of VGG-SS}%
	\label{tab:av_object_detect}
	\begin{threeparttable}
		\begin{tabularx}{\linewidth}{@{}l*{11}{c}}
			\toprule
			Method 			                            &airplane        &baby       	  &cow             &flute           &knife      	 &pigeon     	  &railcar    	   &shaver     	   &ukulele    	    &woodpecker     &$\text{mAP}_{50} \uparrow$ \\
			\midrule\midrule
			Attention \cite{Senocak18}$_{\text{CVPR}18}$&59.2		     &33.2			  &11.4		       &7.1		        &11.9			 &8.3			  &12.6			   &\underline{6.9}&32.5			&4.0		    &21.5 \\
			HardWay \cite{Chen21c}$_{\text{CVPR}21}$   	&\underline{66.7}&28.0			  &9.4		       &\textbf{25.0}	&6.7			 &3.6			  &7.1			   &6.3			   &38.9			&\textbf{10.0}	&22.5 \\
			AVID \cite{Morgado21b}$_{\text{CVPR}21}$    &5.6		     &2.9			  &\textbf{17.5}   &\textbf{25.0}	&10.4			 &9.5			  &14.3			   &0.0			   &30.0			&0.0		    &20.3 \\
			FNAC \cite{Sun23}$_{\text{CVPR}23}$         &44.4		     &\underline{34.6}&10.1		       &\underline{16.7}&5.6			 &3.6			  &\underline{16.8}&1.0			   &\underline{68.0}&\underline{8.9}&\underline{25.8} \\
			SSPL (w/o PCM)   							&50.0		     &25.7		      &8.0		       &\textbf{25.0}	&25.0			 &13.8			  &12.8			   &\textbf{8.3}   &40.0			&8.3		    &25.0 \\               
			SSPL (w/ PCM)								&\textbf{75.6}	 &10.1		      &7.2		       &12.5		    &\textbf{39.5}	 &\underline{16.5}&8.2			   &\underline{6.9}&31.0			&2.9		    &23.8 \\
			SACL       									&\underline{66.7}&\textbf{40.0}	  &\underline{11.5}&2.9		        &\underline{36.1}&\textbf{28.6}	  &\textbf{26.8}   &4.5			   &\textbf{72.2}	&5.0		    &\textbf{26.4} \\
			\bottomrule
		\end{tabularx}%
	\end{threeparttable}
\end{table*}

\subsection{Self-Supervised Object Detection}
We also explore the application of the learned audio-visual representations on self-supervised object detection task. Since the audio-visual similarity map indicates the location of sounding objects, it can naturally be used to derive bounding boxes for training a detector. For all methods, we normalize the similarity map and identify the region where similarity values surpass a predefined threshold (0.7 in all experiments). We then obtain the (pseudo) bounding box by determining the smallest box capable of enclosing this region. To involve the category information in training a class-specific detector, we employ the off-the-shelf self-labeling model \cite{Asano20} to generate (pseudo) class label for each image-audio pair. Finally, a Faster R-CNN \cite{Ren15} detector is trained from scratch on VGG-Sound10k and evaluated on the subset of VGG-SS. Note that for evaluation, we follow prior works in unsupervised visual clustering \cite{Ji19,Van20} and use the Hungarian algorithm \cite{Kuhn55} to match predicted classes with ground truth classes. This matching only occurs after training, ensuring that the detector doesn't rely on any manual labels.

\begin{figure}
	\centering
	\includegraphics[width=\linewidth]{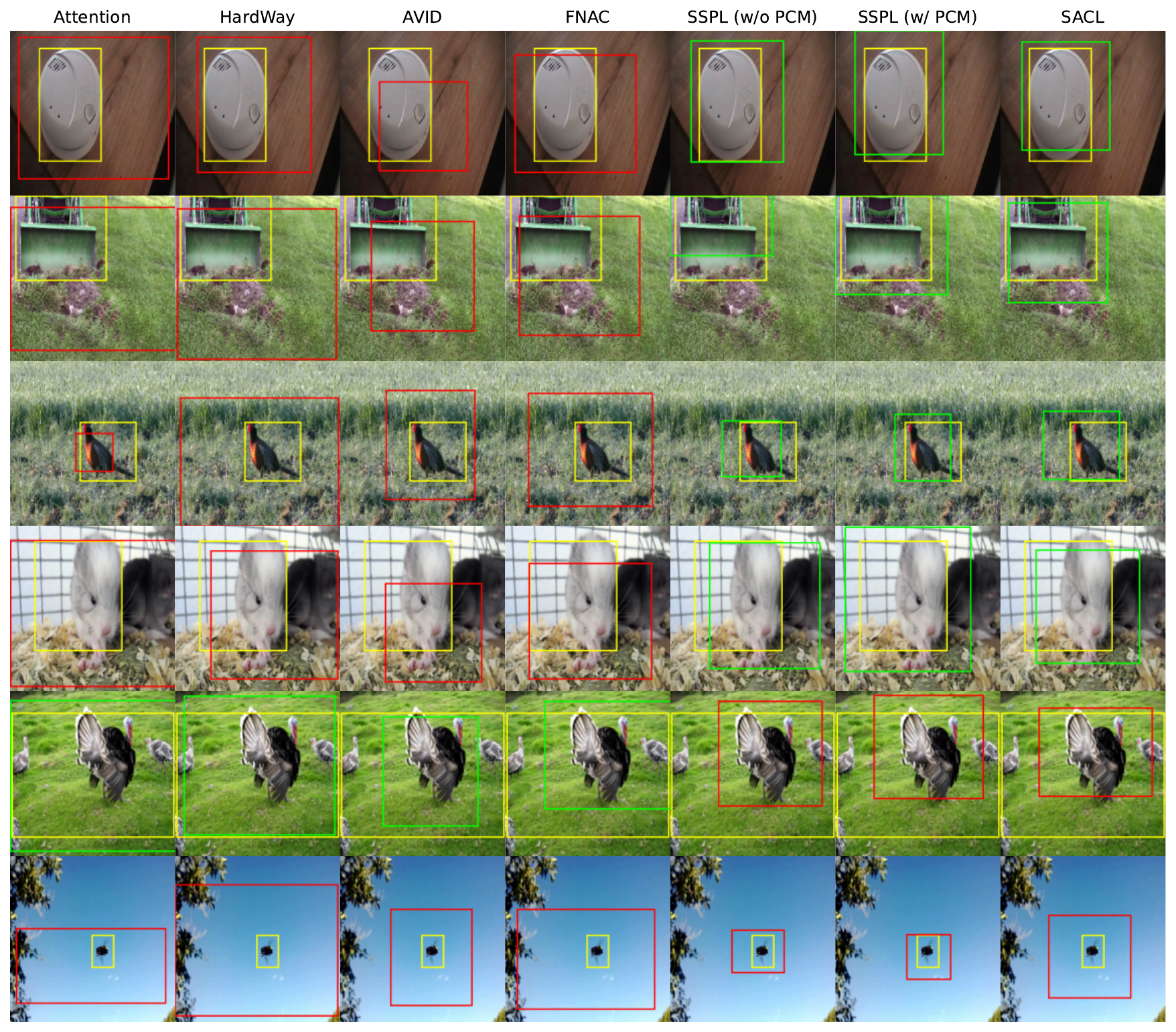}%
	\caption{Visualization of object detection results on the subset of VGG-SS. In each image, only the top scoring box is shown, where a yellow rectangle represents ground truth; a green rectangle denotes successful case ($\text{IoU}\geqslant0.5$); and a red rectangle indicates failure case ($\text{IoU} < 0.5$). The first four rows exhibit instances where our methods excel over the other four methods, achieving higher IoU. By contrast, the fifth row illustrates cases where our methods perform worse than others, resulting in smaller IoU. The last row shows failure examples for all methods.}%
	\label{fig:vis_obj_detect_vggss}
\end{figure}

Table \ref{tab:av_object_detect} reports object detection results. We observe that in general, the methods which function to eliminate false negatives (FNAC, SSPL's, and SACL) work better than those random sampling methods (Attention and HardWay) as for facilitating the detector to find target objects. This manifests that the audio-visual contrastive learning with correct positives and negatives can generate high-quality bounding boxes to train object detectors. Notably, our methods consistently deliver better results for classes of larger objects with distinctive appearances (e.g., airplane in Table \ref{tab:av_object_detect} and tractor in the second row in Fig. \ref{fig:vis_obj_detect_vggss}), but their performance is less robust for smaller objects such as shaver in Table \ref{tab:av_object_detect} and fly in the bottom row in Fig. \ref{fig:vis_obj_detect_vggss}, or for multiple objects appearing simultaneously, like turkeys in the fifth row in Fig. \ref{fig:vis_obj_detect_vggss}. To mitigate these problems, the multiscale approach or more diverse data may be needed during model training.

\section{Conclusion and Future Work}
In this paper, we have presented a novel audio-visual learning framework for sound source localization. The framework is endowed with two alternative learning schemes, SSPL and SACL, to tackle the false negative issue in audio-visual setting. SSPL, as a non-contrastive approach, establishes audio-visual correspondence by mining image-audio positive pairs only. To further lower the positive-pair learning difficulty, we proposed the predictive coding module that enables SSPL to align audio and visual features via cross-modal feature prediction. By contrast, SACL is devised in the contrastive learning scenario, which features the visual feature compaction and false negative detection. By doing so, SACL can effectively find reliable visual anchor and audio negatives, driving the calibrated contrastive learning to benefit from the semantically consistent audio-visual features. Comprehensive experiments have demonstrated our approach's efficacy and expandability with compelling performance, especially achieving the new state-of-the-art on SoundNet-Flickr benchmark.

Overall, this work underlines the importance of reducing ambiguity when sampling negatives in sound localization. We conclude that (1) learning audio-visual correspondence by exploiting positive pairs alone is feasible, provided that additional regularization techniques (e.g., data augmentation, stop-gradient, PCM, \emph{etc}) are adopted. (2) While random negative sampling in contrastive learning hurts performance, it does not mean that all negatives should be discarded. Instead, learning with those audio negatives that are semantically dissimilar with the positive one can boost sound localization, particularly with a high proportion of such negatives within a mini-batch.

Despite the success of our framework, there are still challenges yet to be explored. Given the remarkable performance of SACL with the assistance of image-computable pseudo masks, one challenge is to develop a learnable module for pseudo mask generation \cite{Henaff22}. An end-to-end learning paradigm should have the potential to adaptively segment images, hence bringing simplicity and improving generalization. Another concern is to apply the framework to audio-visual scenarios in the wild, e.g., from single sound source to multiple sound sources \cite{Qian20}. We leave these opening avenues for future work.

\ifCLASSOPTIONcompsoc
  \section*{Acknowledgments}
\else
  \section*{Acknowledgment}
\fi
This work was supported in part by the National Natural Science Foundation of China (Nos. 62276208, 12201490, 12326607, 12371512, and 12301656), in part by the Postdoctoral Fellowship Program of CPSF (No. GZB20230581), in part by the Fundamental Research Funds for the Central Universities (No. xzy012023047), and in part by the Guangdong Basic and Applied Basic Research Foundation (No. 2024A1515010919).

\ifCLASSOPTIONcaptionsoff
  \newpage
\fi

\bibliographystyle{IEEEtran}
\bibliography{IEEEabrv,mybibfile}

\end{document}